%% file: submission.tex
\documentclass[runningheads]{llncs}

\usepackage{graphicx}
\usepackage{amsmath}
\usepackage{amssymb}
\usepackage{booktabs}
\usepackage{caption}
\usepackage[table]{xcolor}
\usepackage[accsupp]{axessibility}  
\usepackage[width=122mm,left=12mm,paperwidth=146mm,height=193mm,top=12mm,paperheight=217mm]{geometry}

\usepackage[pagebackref,breaklinks,colorlinks]{hyperref}

\usepackage[capitalize]{cleveref}
\crefname{section}{Sec.}{Secs.}
\Crefname{section}{Section}{Sections}
\Crefname{table}{Table}{Tables}
\crefname{table}{Tab.}{Tabs.}

\usepackage{comment}

\input{macros.tex}

\begin{document}
\pagestyle{headings}
\mainmatter
\def\ECCVSubNumber{5789}  

\title{Inpainting at Modern Camera Resolution by Guided PatchMatch with Auto-Curation}


\titlerunning{SuperCAF}
%
\author{Lingzhi Zhang\inst{1} \and Connelly Barnes\inst{2} \and Kevin Wampler\inst{2} \and Sohrab Amirghodsi\inst{2} \and \\ Eli Shechtman\inst{2} \and Zhe Lin\inst{2} \and Jianbo Shi\inst{1}}

\authorrunning{Zhang et al.}
%
\institute{University of Pennsylvania \and
Adobe Research \\
}
\input{figs/teaser.tex}

\begin{abstract}
Recently, deep models have established SOTA performance for low-resolution image inpainting, but they lack fidelity at resolutions associated with modern cameras such as 4K or more, and for large holes. We contribute an inpainting benchmark dataset of photos at 4K and above representative of modern sensors. We demonstrate a novel framework that combines deep learning and traditional methods. We use an existing deep inpainting model LaMa~\cite{suvorov2021resolution} to fill the hole plausibly, establish three guide images consisting of structure, segmentation, depth, and apply a multiply-guided PatchMatch~\cite{barnes2009patchmatch} to produce eight candidate upsampled inpainted images. Next, we feed all candidate inpaintings through a novel curation module that chooses a good inpainting by column summation on an 8x8 antisymmetric pairwise preference matrix. Our framework's results are overwhelmingly preferred by users over 8 strong baselines, with improvements of quantitative metrics up to \emph{7.4 times} over the best baseline LaMa, and our technique when paired with 4 different SOTA inpainting backbones improves each such that ours is overwhelmingly preferred by users over a strong super-res baseline.

\end{abstract}


\section{Introduction}
\input{figs/overview}

Image inpainting involves removing a region and replacing it with new pixels so the modified photo is visually plausible. We develop a method that can take an off-the-shelf low-res inpainting deep model and extend it to modern camera resolutions. Applied to LaMa~\cite{suvorov2021resolution}, this yields a new a SOTA method for inpainting at modern camera resolutions that dramatically outperforms all existing models.

Traditional patch-based synthesis approaches, such as Wexler~et al. \shortcite{wexler2007space}, Barnes~et~al.~\shortcite{barnes2009patchmatch}, and Darabi~et~al.~\shortcite{darabi2012image} were used for high-quality image inpainting at arbitrary resolutions. Recently, the state-of-the-art for \emph{low resolution} image inpainting has been advanced by deep convolutional methods, such as Zeng~et~al.~\shortcite{zeng2020high}, Zhao~et~al.~\shortcite{zhao2021comodgan}, and LaMa~\cite{suvorov2021resolution}.
When we compared these methods, we noticed an interesting trade-off. Patch-based methods such as PatchMatch~\cite{barnes2009patchmatch} can synthesize high-quality texture at arbitrarily high resolution but often make mistakes regarding structure and semantics. On the other hand, deep convolutional methods tend to generate inpaintings with plausible structure and semantics but less realistic textures.  Worse still, most deep  methods are limited to output resolutions such as 512 or 1024 pixels image along the max dimension\footnote{With the exception of HiFill~\cite{yi2020contextual} and LaMa~\cite{suvorov2021resolution}, which we discuss in related work.}, \emph{much smaller} than modern camera resolutions which are typically 4K or above.

Based on our observations, we first collected a dataset of 1045 high-quality photos at 4K resolution or above representative of modern sensors and paired them with freeform and object-shaped hole masks. This dataset has approximately \emph{2 orders of magnitude} more pixels per image than the dataset Places2 \cite{zhou2017places} that is commonly used for deep inpainting evaluation. Interestingly, photos in the Places2 dataset often used for evaluation have fewer pixels than the world's first digital camera~\cite{fujix}, the FUJIX DS-1P, produced in 1988, so common evaluation practices are \textbf{more than 3 decades} behind sensor technology. We encourage the community to evaluate on our dataset, since datasets such as Places2 \cite{zhou2017places} are \emph{not representative} of modern camera sensors, and which may therefore mislead researchers by giving improper guidance to the community.

After establishing the dataset, we next married the complementary advantages of patch-based synthesis in generating high texture quality at arbitrarily high resolution and deep networks, which can better predict structure and semantics. Our hybrid pipeline works as illustrated in  \fig{fig:overview}. 
We first inpaint the hole region using an off-the-shelf deep network, such as LaMa~\cite{suvorov2021resolution}, at 512 pixels along the long edge of the image. This method typically establishes a reasonable semantic layout and structure inside the hole. Still, as can be seen in \fig{fig:teaser}, the texture quality is often poor if we zoom in. Next, we extract several guide images using existing methods: depth~\cite{yin2021CVPR}, structure~\cite{xu2012structure}, and panoptic segmentation~\cite{li2021fully}. We then use these guides in a multi-guided PatchMatch~\cite{barnes2009patchmatch} implementation to perform patch-based image inpainting. Depending on the input photo, the best result may be obtained by different combinations of guides, so we produce a set of eight guided PatchMatch \emph{candidate inpainting} results using multiple combinations of guides.

Besides the novelty of our overall framework, our key technical innovation is creating a \textbf{novel automatic curation module} whose architecture is shown in \fig{fig:curation_module}(a). This curation module automatically selects one good inpainted image out of the 8 candidates by an architecture suited for making subtle comparisons and contrasts between similar images. It does this by constructing an antisymmetric 8x8 matrix whose entries are populated in pairs by a network that learns to predict \emph{antisymmetric pairwise human preferences} for every possible pair of candidate inpaintings. The preferred inpainted image is selected by taking the row with max column sum for the 8x8 matrix. Different from previous work such as RealismCNN~\cite{zhu2015learning} and LPIPS~\cite{Zhang_2018_CVPR} and image quality assessment papers, our network estimates pairs of entries $M_{ij}$ and $M_{ji}$ at a time for an 8x8 matrix $M$ that is \emph{antisymmetric} i.e. $M = -M^T$, and the pairwise structure is critical because it allows the network to differentiate between subtle feature differences. On pairs of images, our curation has near-human performance. 

We conducted quantitative experiments with metrics commonly used to evaluate deep networks and multiple user studies for our evaluation. Our method dramatically outperforms state-of-the-art deep inpainting methods according to our quantitative metrics and user studies. User preferences are between 79\% and 96\% for our method, \emph{every one of our 7} user study results in Tables \ref{tab:comp_with_sota}-\ref{tab:pm_random_ours} has statistically significant preference for ours, and quantitative metrics improve by up to 7.4  times over strong baselines like LaMa~\cite{suvorov2021resolution} coupled with Real-ESRGAN~\cite{wang2021real}. Because our method uses patch-based synthesis, it can easily \emph{scale to arbitrary resolution}: in the supplemental, we show results on images up to 62 megapixels. We show in our experiments that our method can be combined with four different deep inpainting baselines and improves \textbf{every} one of them for modern camera resolution inpainting according to quantitative metrics and user studies. Most convolutional models cannot scale to modern camera resolutions, except HiFill~\cite{yi2020contextual} and LaMa~\cite{suvorov2021resolution}, which our evaluation shows we outperform.

Our paper contributes: (1) A benchmark dataset of 4K+ resolution images with holes appropriate for evaluating inpainting methods as they perform on modern camera sensors and evaluation of 9 methods on this dataset; (2) A novel high-resolution inpainting framework which shows that deep inpainting models need not give high quality results only on low resolutions. This framework has a choice of guides that perform well, which were nontrivial to choose and required extensive empirical investigations as discussed in \sect{sec:guides_for_PatchMatch}, and is the first to explore combinations of multiple guides with a mechanism to automatically choose between them; (3) A curation module (seen for the first time in inpainting) with near-human performance that chooses a good inpainted image in a new way by populating an 8x8 antisymmetric preference matrix by comparing images one pair at a time and column-summing that matrix; (4) State-of-the-art results on the natural image inpainting task from both quantitative and user studies, with dramatic improvements over 8 strong baselines; (5) Our method can be combined with a variety of SOTA deep inpainting baselines and improves each of them for modern camera resolution inpainting. To enable reproducibility, we plan to release our full benchmark dataset including results of all methods and curation module test code and weights.

\vspace{-10pt}
\section{Related Work}
\vspace{-10pt}

\textbf{Patch-based synthesis and inpainting.} Our approach uses the patch-based inpainting method of Wexler~et~al.~\shortcite{wexler2007space} as implemented in the PatchMatch framework~\cite{barnes2009patchmatch}. Inspired by image analogies~\cite{hertzmann2001image}, we added to this basic inpainting framework multiple guiding channels. Similar guided texture synthesis approaches have been used for stylizing renderings~\cite{fivser2016stylit} and stylizing video by example~\cite{benard2013stylizing,jamrivska2019stylizing}, but the choice of guides is \emph{nontrivial and application-dependent}: we performed many empirical investigations of alternative options to finally choose the guides used in this paper, as we discuss later in \sect{sec:guides_for_PatchMatch}. Image Melding~\cite{darabi2012image} improved patch-based synthesis results with geometric and photometric transformations and other means. We incorporated a gain and bias term from that paper to obtain better inpaintings for regions with smooth gradient changes in intensity. Kaspar~et~al.~\shortcite{kaspar2015self} performed texture synthesis using a patch-based optimization with guidance channels that preserve large-scale structures. Several papers explored initialization and search space constraints that use scene-level information that is different from ours~\cite{he2012statistics,huang2014image,diamanti2015synthesis}.


Recently, some papers have explored learning good features by making PatchMatch differentiable~\cite{duggal2019deeppruner,zhou2021cocosnet}. Because of our modern camera resolution photos, multi-res and many iterations of filling used in PatchMatch ~\cite{barnes2009patchmatch}, and because differentiable PatchMatch techniques track multiple candidate patches, the GPU memory requirements of applying a differentiable PatchMatch naively in our context are far beyond the capacity of today's GPUs. Thus, we use a non-differentiable PatchMatch with off-the-shelf guide features, and rely on the downstream curation module to pick good guide features.


\textbf{Neural network image inpainting.} One significant advantage of convolutional neural network inpainting methods is that they can gain some understanding of semantic information such as global and local context, edges, and regions. Context encoders used a CNN to fill in a hole~\cite{pathak2016context} by mapping an image with a square hole to a filled image. Iizuka~et~al.~\shortcite{iizuka2017globally} used two  discriminators to encourage the global and local appearance to both be reasonable. Yu~et~al.~\shortcite{yu2018generative} proposed a contextual attention model which can effectively copy patches from outside the hole to inside the hole, however, it is limited in resolution because of the brute-force dense attention mechanism. Liu~et~al.~\shortcite{liu2018image} and Yu~et~al.~\shortcite{yu2019free} improved inpainting results by masking and gating convolutions, respectively. Zhao~et~al.~\shortcite{zhao2021comodgan} better address the situation of large holes by co-modulating the generative decoder using both a conditional and stochastic style representation.

A number of papers recently attempt to separate the scene in terms of edges~\cite{nazeri2019edgeconnect,xiong2019foreground}, structure-texture separation, such as Ren~et~al.~\cite{ren2019structureflow} which uses relative total variation (RTV)~\cite{xu2012structure}, or Liu~et~al.~ \cite{liu2020rethinking}. Our approach is inspired by these works and uses structure from RTV~\cite{xu2012structure} and a panoptic segmentation map to guide PatchMatch. 


A few recent works attempt to train neural networks that can output higher resolution results. ProFill~\shortcite{zeng2020high} uses a guided upsampling network at a resolution of up to 1K on the long edge. HiFill~\shortcite{yi2020contextual} introduced a contextual residual aggregation module that weights residuals similar to those in a Laplacian pyramid ~\cite{burt1987laplacian}, at up to 8K resolution, but according to our evaluation its results are worse than our method. The work LaMa~\cite{suvorov2021resolution} was trained on 256x256 patches but can be evaluated on images up to 2K.


\textbf{Visual realism.} For pretraining our curation network, we use a similar idea as  RealismCNN~\cite{zhu2015learning}, which learns a visual realism score for composite images using a large dataset of images and a synthetic compositing pipeline. In our case, however, we generate millions of fake inpainted images using our pipeline, and pretrain a network to classify these images as fake and real images as real. Inspired by LPIPS~\cite{Zhang_2018_CVPR} and image quality assessment (IQA) papers (e.g.,~\cite{bosse2017deep,talebi2018nima,zhu2020metaiqa}), we then fine-tune on real human preferences between pairs of synthetic inpainted images, but we use a  different architecture and inference that involves predicting entries one pair at a time in an \emph{antisymmetric matrix} and column-summing it. In contrast, LPIPS~\cite{Zhang_2018_CVPR} learns a \emph{symmetric} distance metric, full-reference IQA papers typically are also estimating some perceptual distance that is not antisymmetric, and no-reference IQA papers use a single image as input. See \fig{fig:curation_module} for an illustration. Also related is  Wang~et~al.~\shortcite{wang2020cnn}, which trains a ``universal" detector that can distinguish between CNN-generated images and real images. 

\vspace{-10pt}
\section{Inpainting Dataset}
\label{sec:datasets}
\vspace{-5pt}

To benchmark the performance of our method and 8 other methods, we collect 1045 high-quality images at modern sensor resolutions from two sources: DIV8K \cite{gu2019div8k}, and a test set portion of the dataset of photos taken by the authors and their collaborators mentioned earlier in \sect{sec:curationnetwork}. The photos span a diverse variety of scenes including indoors and outdoor urban photography including many architectural styles, nature and wildlife, macro photos, and human subject photos. The photos are all at least 4K pixels along the long edge. The mean megapixel count is 20 and the maximum is 45, excluding high-res panos that go up to 62 MP. DIV8K \cite{gu2019div8k} has previously been used for super-resolution tasks and contains images with resolutions up to 8K. We chose all 583 images in DIV8K that were above 4K resolution on the long axis. 


For each test image, with equal probability, we sample either a free-form mask or an instance mask using the same process and hole dataset as in ProFill \cite{zeng2020high}. Different from ProFill, because our photos are from modern sensors, we generate larger holes with a mean hole size of 2.3 MP: see the supplemental for details. 
No test images are seen during training. To enable reproducibility, we plan to release the testing dataset and results for all methods. Experimental results are shown later in \sect{sec:comparisons}. 

\section{Our Hybrid Synthesis Method}

\subsection{Multi-guided PatchMatch for Image Inpainting}

PatchMatch~\cite{barnes2009patchmatch} is an efficient randomized correspondence algorithm that is commonly used for patch-based synthesis for images, videos~\cite{benard2013stylizing,jamrivska2019stylizing} and neural feature maps~\cite{liao2017visual,zhou2021full}. 
For image synthesis, PatchMatch repeatedly performs matching from the region being synthesized to a reference region: in our case, the matching is from the hole to the background region. We implement the method of Wexler~et~al.~\shortcite{wexler2007space}, with default hyperparameters for PatchMatch (e.g. 7x7 patches) and Wexler~et~al. One key advantage of PatchMatch is that it can efficiently scale to arbitrary resolutions while preserving high-texture fidelity. We extend this basic method in two ways: we allow multiple guides to be used (as in~\cite{hertzmann2001image}), and we implement the gain and bias term from Image Melding~\cite{darabi2012image}.

For the multiple guides, we modify the sum of squared differences (SSD) used in Equation (5) of PatchMatch.  Instead of computing SSD over a 3 channel color image, we compute a weighted SSD over a $3+m$ channel image with channel weight $w_i$, where the first 3 channels are RGB and remaining channels are guides. 
\begin{equation}
    w_i = \begin{cases}
w_c/3 &i \leq 3\\
(1-w_c)/m & i > 3
\end{cases}
\end{equation}
Using a separate validation set of images and manual inspection, we tried different settings of $w_c$ and empirically found the best setting is $w_c = 0.6$ if there is no structure guide, otherwise $w_c = 0.3$. Structure and RGB information are highly correlated so we decrease the RGB weight if there is a structure guide.

We implement the gain and bias term from Image Melding~\cite{darabi2012image} because we find it helps improve inpainting quality in cases where there are subtle gradients, such as in the sky. \eccvorarxiv{We implement this term using RGB color space.}{We implement this term the same way as described in the ``Search" portion of their Section 3.1 but using RGB color space.} \arxiv{We use the same min and max for gain as in Image Melding.  For the bias, because we are using RGB color space, we obtained the best quality with a min and max of $[-0.05, 0.05]$ times the maximum channel value (255).}

\subsection{Guides Used for PatchMatch}
\label{sec:guides_for_PatchMatch}


We performed extensive investigations of many possible guides, and empirically settled on the three selected in this paper --- structure, depth, and panoptic segmentation --- because they worked the best on our validation set. Although our main novel low-level technical contribution is our curation module that we describe in the next section, the existence of appropriate guides, the empirical work needed to choose them and our overall novel multi-guided inpainting framework with curation also forms a high-level technical contribution.

Our pipeline works by first inpainting the hole using an off-the-shelf deep network LaMa~\cite{suvorov2021resolution} at 512 pixels on the long edge and then extracting three guide images: depth, segmentation, and structure. Depth is a useful cue as regions at a similar distance from the camera have usually more relevant content. This is especially important for slanted surfaces (e.g. the ground) where scale and focus properties of texture vary with depth. We also often want to sample semantically similar content, motivating the segmentation guide. However, the segmentation labeling might not be fine-grained enough to distinguish between different types of floor tiles or wall colors. Therefore we add the structure guide that captures the main edges and color regions in the image, and abstracts away the texture.

For depth prediction, we used a recent method by Yin~et~al.~\shortcite{yin2021CVPR}, which is retrained using a DPT~\cite{ranftl2021vision} backbone to obtain better results. For panoptic segmentation prediction, we used PanopticFCN~\cite{li2021fully} retrained with a ResNeSt~\cite{zhang2020resnest} backbone, which obtains a higher panoptic quality of 47.6 on the COCO val set. For structure prediction, we extract the structure using RTV~\cite{xu2012structure}, that was shown effective for inpainting by Ren~et~al.~\shortcite{ren2019structureflow}.

We performed extensive investigations of many alternative possible guides, and choose these three because they gave the best quality. We tried raw RGB color instead of structure, but this produced worse results due to patch matching becoming less flexible (see supplemental for details), Iterative Least Squares texture smoothing~\shortcite{liu2020real}, but the latter had color bleeding artifacts that caused worse inpaintings, and Gaussian blurred RGB color and bilateral filtered RGB instead of structure, but these produced artifacts in the patch synthesis. \eccvorarxiv{We also tried different segmentation and depth modules and settled on the above ones because they gave better patch synthesis results.}{We also tried semantic segmentation with HRNetV2~\cite{sun2019high} trained on ADE20k~\cite{zhou2017scene} but switched to a more recent panoptic model because it resulted in more accurate boundaries and avoided copying patches between instances. For depth, we initially used Yin~et~al.'s~\shortcite{yin2021CVPR} model but switched to the DPT backbone because it produced more accurate results. For the DPT backbone, we tried depth, disparity, and log-depth, and settled on log-depth because it produced better results due to better separating between near, mid, and far regions.}


Is there a universal `best' guide for all images? We find that different choices of guides may give the best final inpainting result, depending on the input image and hole. 
Empirically, we find that the structure guide can help PatchMatch find patches with consistent edges and fine-scale structures, depth guides can be handy when the images have gradually changing depth such as outdoor photography, and segmentation guides can be useful when the segmentation map is accurate for preventing patches from leaking into the wrong semantic or object region. In general, we find no single guidance map by itself is sufficient to get the best quality of results, and multiple guidance maps are needed. In many applications it is desirable to have a fully automatic image inpainting process and we found that a fixed weighted combination of guides leads to sub-optimal results. Thus, as shown in \fig{fig:overview}, we generate a variety of results using different guides and use our curation module to automatically choose a good one. Specifically, we use a simple scheme where we can either enable or disable each of our 3 guides, so the total number of possibilities for guides are $2^3 = 8$, including the use of no guides, and then we select among those eight generated results. We show an example of how the guides influence the results in the supplemental.



\subsection{Curation Module}
\label{sec:curationnetwork}

Suppose you are given two inpainted candidate images and asked, \emph{``Which of these do you prefer, or do you have no preference?"} We were inspired by how humans carry out this task: since the images usually look similar, we carefully compare and contrast  between subtle differences in visual features to determine which image is slightly better overall. Our architecture thus is designed to enable such \emph{subtle comparisons and contrasts} between features within inpainted images. In \fig{fig:curation_module}, we show an architecture diagram of our novel curation module and a comparison with other common architectures such as LPIPS~\cite{Zhang_2018_CVPR} and no-reference IQA. Notably, our curation module has a  different architecture and inference methodology that populates entries in a matrix $M$ that is \emph{antisymmetric} due to the paired preference task and column-sums that matrix to determine the relative preference vector of one inpainting candidate as contrasted with others.

\input{figs/curation_module}

\textbf{Pretraining.} Inspired by RealismCNN~\cite{zhu2015learning}, our curation network backbone $F$ is first pretrained to classify for a given image, whether it is a real image or a fake inpainted image as output by our pipeline. Our reasoning is that we observed that the initial pretrained network predictions have  correlation with human perception of inpainting quality, and can allow the network to learn good features over a very large number of photos, but we need to later fine-tune on human preferences to obtain performance close to humans. 
We generated a dataset for pretraining the curation network by collecting 48229 diverse photos that are at least 2K in resolution, generating 10 synthetic holes for each, and then generating 8 guided PatchMatch results. This results in more than 3 million inpainted images at 2K resolution. 
Our curation network backbone $F$ is EfficientNet-B0~\cite{tan2019efficientnet}. We modify the input to take 4 input channels and pretrain the network from scratch. We trained for one epoch using a binary cross-entropy loss, after which we obtained train, test, val accuracies of 98.9\%, 99.3\%, 99.2\%, respectively. Please see the supplemental for more details. 


\textbf{Fine-tuning on human preferences.} In nearly all cases, the pretrained curation network $F$ can easily distinguish between real and our inpainted results, but it was not specifically trained to predict human preference among different inpainted results. Therefore, we next fine-tune our network for a paired preference task. By subsampling the dataset described in the last subsection, and comparing sampled pairs of the 8 guided PatchMatch results against the others, we gather approximately 33000 synthetic inpainted image pairs for which we gather human preferences. We discuss in our supplemental lessons learned in gathering these preferences. For each image pair, our model works by featurizing each of the images in the pair through a shared-weight pretrained EfficientNet backbone $F$, and then using a small MLP to predict 3 classes that the human preference data contain: prefer left image, tie, prefer right image. 

In contrast to perceptual distances such as LPIPS~\cite{Zhang_2018_CVPR}, we have a different task where our model predicts an antisymmetric preference. In particular, if one swaps the left and right image, one would expect the preference for left or right image to also swap. We thus impose this swapping as a data augmentation, by doubling each original batch to include a swapped copy of the batch: we found this accelerates and stabilizes training. We include a variety of standard augmentations that we list in the supplemental.

\textbf{Inference for curation.} Our network is trained on paired preferences, but at inference time, we want to compare 8 guided PatchMatch inpaintings and establish a preference for each, and a preferred ordering. Moreover, in alternative implementations, one might wish to compare more or fewer images. Thus, to compare $n$ inpainted candidate images, we form an $n \times n$ matrix $M$, where $M_{ij}$ is the probability of preferring method $i$ over method $j$. We establish this probability for all pairs $i, j$ with $i<j$ by setting $o_{ij}^{(k)}$ for $k=1,2,3$ as the 3 softmaxed outputs of the MLP for the pair, and then compute $M_{ij} = o_{ij}^{(3)} - o_{ij}^{(1)}$ and $M_{ji} = -M_{ij}$. The ground truth and  prediction are antisymmetric matrices i.e. $M = -M^T$. The preference of a given inpainted image $i$ \emph{in the context of the other images} is the average of row $i$ of $M$. In this way, we recover the same antisymmetric paired preferences in the special case of $n=2$, but also generalize to establishing preferences among arbitrary numbers of images.


The input resolution for our curation network is 512x512, however, the photographs to be inpainted can have 1 to 2 orders of magnitude more pixels. We use an operation called ``auto crop" to resize a crop region around the hole that contains approximately 25\% hole pixels and surrounding context  to the target resolution. Please see the supplemental for details about  automatic cropping.
\section{Experiments}



\subsection{Curation Module}
\label{sec:curation_experiments}

We show in \tbl{tab:curation_v2} the performance on human paired preference data for our curation module, ablations, and comparisons. The table is computed from the human preference dataset previously discussed in \sect{sec:curationnetwork}. In our user study, if a human expresses a preference for one image, we ask if the preference is strong or weak. Because the task itself is challenging, we also report \textbf{easy cases} as those where mean human preference is strongly for one image.

\input{tables/curation_v2}

We list our network and human performance in \tbl{tab:curation_v2}. Human preference is determined by collecting an additional opinion for a random subset of images. Our curation module outperforms all other alternatives and is only slightly worse than human performance by both metrics. Although the accuracy for both humans and ours is ``only" 56-57\%, this is already much better than random chance at 33\%, and this is because of the difficulty of the task, where for many fills it is hard to tell whether they are tied in quality or one or the other is preferred. For easy or unambiguous cases, the accuracies for humans and ours are both much better, at 85-86\%. Since models can overfit to the training set, we always report the checkpoint with highest validation accuracy.

We next discuss the ablations in \tbl{tab:curation_v2}. ``Ours No Pretraining" skips the pretraining, which is necessary for best generalization. ``Ours Fewer Augmentations" is an ablation where removing JPEG compression, rotation, and noise reduces accuracy. For ``Ours No Mask," we do not input the hole mask. ``Ours Late Fusion Variant" modifies the pretraining so  instead of using one classifier, both real and fake image are featurized with a shared-weight EfficientNet backbone and compared with an additional MLP. ``Ours Early Fusion" modifies the network by concatenating both images with mask and feeding this through a single EfficientNet backbone followed by MLP. 

We find that fine-tuning only the MLP for the human preference task and freezing the backbone network weights is insufficient. ``Ours Freeze Backbone" freezes backbone weights after pretraining and only fine-tunes the MLP. Similarly, ``NIMA\cite{talebi2018nima} w/ MLP" and ``MetaIQA+\cite{zhu2021generalizable} w/ MLP" use  pretrained, frozen no-reference image quality assessment backbones, and fine-tune the MLP.

Our \tbl{tab:pm_random_ours} shows that the guided inpainting chosen by our curation module out of all 8 options outperforms both a random choice of a guided fill and Photoshop's Content-Aware Fill: the outperformance in user preference is particularly strong. That table is for the inpainting dataset, which is described next. \arxiv{Please see our supplemental document for additional experiments involving the curation module.}

\subsection{Comparison with Other Methods}
\label{sec:comparisons}







We compare our algorithm with eight state-of-the-art image inpainting methods, quantitatively and qualitatively. 
Among all these methods, HiFill \cite{yi2020contextual} can run on images up to 8K resolution like our method, the work LaMa~\cite{suvorov2021resolution} states that they can generalize to higher resolutions up to around 2K, ProFill \cite{zeng2020high} can run on images up to 1K resolution, and the rest of the methods can only run on images up to 512 x 512.

\input{tables/comp_with_sota}
\input{tables/improve_4_deep_and_pm_random_ours}

The main focus of our method is inpainting of high resolution images of size 4K and beyond. We thus generate full resolution images for all methods. We attempted to make the comparison as generous as possible for baseline methods by applying Real-ESRGAN~\cite{wang2021real} for super-resolution to increase all methods with limited output resolution back to the native image resolution. We chose Real-ESRGAN~\cite{wang2021real}, since it is the state-of-the-art SR algorithm for real-world images and is robust to visual artifacts in the input. We apply HiFill~\cite{yi2020contextual} at native resolution, and ProFill~\cite{zeng2020high} at its maximum of 1K resolution. For LaMa~\cite{suvorov2021resolution}, we found that although it can be applied at higher resolutions, we obtain best quality by applying it at a resolution where the maximum axis is 512 pixels. Additionally, we tried two scenarios for the baselines: running on the auto-crop region discussed earlier in \sect{sec:curationnetwork} and running on the full image. In our evaluation we found no advantage for the baselines run on the auto-crop region so we used the full image scenario.\arxiv{ See our supplemental document for more experiments and details.}



For the quantitative evaluation, we evaluated six popular metrics: Peak Signal-to-Noise Ratio (PSNR), SSIM~\cite{wang2004image}, LPIPS \cite{zhang2018unreasonable}, the recently improved version of FID \cite{parmar2021buggy}, and P-IDS and U-IDS from CoModGAN~\cite{zhao2021comodgan}, which were recently shown to correlate highly with human perception. Because the holes are very large (2.3 megapixel on average) and valid fills can have quite diverse contents that differ greatly from the original image, we feel the metrics \emph{FID and P/U-IDS are most appropriate}, and we also show LPIPS, and we report PSNR and SSIM in the supplemental. The quantitative results are shown in \tbl{tab:comp_with_sota}. For P/U-IDS, because of the dataset size of 1045 images, we also apply vertical and horizontal augmentation for the full image to improve convergence of those metrics.

In \tbl{tab:comp_with_sota}, we report quantitative metrics for two scenarios: ``full" indicates a square crop region around the entire inpainted region was used, as determined by auto-crop~\sect{sec:curationnetwork}, and ``patch" indicates 10 smaller randomly sampled 256x256 crop regions drawn at consistent position from locations where the patch center is a hole pixel. We note that ours outperforms all baselines by the metrics in the table. We particularly note that in the patch scenario, for the metrics FID, P-IDS, and U-IDS, we \emph{dramatically} outperform the SOTA method LaMa~\cite{suvorov2021resolution} by factors \emph{between 2.3 to 7.4 times}, and the outperformance can be even greater for other methods. This is because our method has much higher texture fidelity at the finest resolutions, since it can copy relevant background patches via PatchMatch. These textures form a coherent whole, as indicated by  outperformance via other metrics.

As often mentioned in the inpainting literature (e.g.,~\cite{yu2018generative,yu2019free}), image inpainting lacks a good quantitative evaluation metric, and there is no single metric that can be used to gauge real image quality. Thus, we also conduct two user studies for randomly sampled 200 test images in comparison with the top four methods using Amazon Mechanical Turk. For the first user study, we use the ``full" image scenario described earlier, and in the second user study, we use a ``boundary patch" setup. The boundary patch is determined by randomly sampling a 512x512 crop region, centered on a boundary pixel of the hole region. This setup allows human participants to easily contrast the texture synthesized inside the hole with that in the background region, and thus assess the suitability of methods for inpainting at modern camera resolutions. In the user study, we give user output images from all methods with randomized order, and ask users to pick the best image for each case. For each of the two user studies, we recruited more than 150 users and asked each user to evaluate a randomly sampled batch of 20 images from the whole test set. 
The study results are shown in the last two columns of \tbl{tab:comp_with_sota}. In the ``full" scenario, our automatically selected guided PatchMatch outperforms the other methods by a large margin of \emph{4.6 times to 21 times}, and in the ``boundary patch" scenario, our method outperforms alternatives even more strongly, by \emph{6 times to 11 times}. \eccvorarxiv{Users in general very strongly prefer our method for inpainting at modern camera resolutions, but in the crop scenario where a user focuses on high-resolution detail --- as might be important for large displays or large format prints --- our method performs better still.}{This indicates that users in general very strongly prefer our method for inpainting at modern camera resolutions, but in the crop scenario where a user focuses on high-resolution detail --- as might be important for large displays or large format prints --- our method performs even better still.}

In \tbl{tab:improve_4_deep_models}, we show our method can be used in combination with 4 different baseline methods that perform  the initial coarse-scale inpainting: ProFill\cite{zeng2020high}, CoModGAN~\cite{zhao2021comodgan}, MADF~\cite{zhu2021image}, and LaMa~\cite{suvorov2021resolution}. In every case, our method outperforms the baseline with Real-ESRGAN~\cite{wang2021real} super-resolution applied. We ran user study containing pairs of images, and found the user study preferences strongly prefer our method, with between \emph{4.5 and 8.5 times} higher preference for ours over the baseline. Our method is suitable when combined with a variety of deep inpainting baselines, and greatly improves user preference over alternatives. We show similar results for 3 older inpainting models in the supplemental.

In \tbl{tab:pm_random_ours}, we compare our method to two other baselines: Photoshop's Content-Aware Fill (CAF), which is based on PatchMatch~\cite{barnes2009patchmatch}, and a baseline that randomly picks with equal probability one of our eight guided PatchMatch results. Our method again performs best for the quantitative metrics. We ran a user study on 100 images where all methods have different outputs, and again find our method is \emph{strongly preferred 3.4 to 4 times} more than the other two baselines. This indicates that our method outperforms a strong commerical baseline of Photoshop's CAF, which is used by professionals to manipulate photos at modern camera resolutions, and shows that our curation outperforms a random guided PatchMatch result. We include many photographic results in our supplemental material, and show that the preference for ours is statistically significant for \emph{all} 7 of the above user studies.

\input{figs/qualitative_results}



\subsection{Running Time}

On a 3.6 GHz 8 core Intel i9-9900K with 11 GB NVidia RTX 2080 Ti, for a representative 12 MP image with 4K resolution, \eccvorarxiv{our naive implementation takes 23.0 seconds, and our optimized implementation takes 2.5 seconds by initially computing PatchMatch results at 1K, running curation, then using another PatchMatch to obtain the 4K result.}{model inference times are 0.065 s for LaMa at 512 res, 0.084 s for depth, 0.162 s for segmentation, and 0.101 s to run our curation module on all 8 PatchMatch results. Structure map inference is 0.21 s using our C++ implementation and  PatchMatch takes 22.35 s to generate all 8 results at 4K. The sum of these times is 23.0 s. We also have an optimized implementation that gives similar quality, which reduces inference time to 2.5 s for 4K images by initially computing PatchMatch results at 1K, running curation in one batch on a NVidia RTX 3080, and using another PatchMatch to obtain the 4K result. We use the non-optimized implementation in our results.} 

\section{Discussion, Limitations, Future Work}
\label{sec:limitations}


\eccvorarxiv{
Our method has some limitations, which could be mitigated through user interactions such as manually picking guides. Generally, PatchMatch is good at synthesizing texture and repetitive regular structures, but structures under perspective transformations in architecture can be broken. We occasionally observe small amounts of blur especially near the hole boundary: this might be addressed by using curation in a smarter way such as an iterative fill~\cite{zeng2020high,Zhang_2022_inpaint_artifacts}. Occasionally, we observe repetitions of salient patches: these might be mitigated by incorporating patch usage budgets~\cite{fivser2016stylit,kaspar2015self} combined with saliency. GANs may produce amazing results by hallucinating content not present in the input image, but they can also hallucinate bizarre artifacts. We use patch-based synthesis throughout the image, however, patch-based synthesis can remove unique features, so the result could be allowed to deviate from patch-based synthesis if we believe a hallucinated output is a good one.

}{
Our method has some limitations, which could be mitigated through user interactions such as manually picking guides or improved upon by future work. Generally, PatchMatch is good at synthesizing texture and repetitive regular structures, but structures under perspective transformations particularly in architecture can be broken: this might be addressed by using special guidance~\shortcite{huang2014image}. We occasionally observe small amounts of blur especially near the hole boundary: we find our curation network avoids blur artifacts, so this might be addressed by using curation in a smarter way such as an iterative fill~\cite{zeng2020high}. Occasionally, we observe repetitions of salient patches: these might be mitigated by incorporating patch usage budgets~\cite{fivser2016stylit,kaspar2015self} combined with saliency. GANs may produce amazing results by hallucinating content not present in the input image, but they can also hallucinate bizarre artifacts. We use patch-based synthesis throughout the image, however, patch-based synthesis can remove unique features, so the result could be allowed to deviate from patch-based synthesis if we believe a hallucinated output is a good one. We show these limitations in the supplemental.

Given the dramatic improvements in quantitative metrics and user studies, we wanted to give two speculative takeaways: (1) Future advances in deep modules such as inpainting could benefit hybrid approaches such as ours; (2) We speculate that capacity limits in deep models may in the near- and mid-term pose a challenge for non-exemplar-based methods to produce high fidelity texture at modern sensor resolutions. Thus, hybrid approaches like ours or deep exemplar-based approaches may offer a road forward for related problems in image and video, and our guides could serve as a starting point for such research.
One potential negative societal impact of our work is generating fake photos, i.e. removing an important person or object, and delivering misinformation to the public for malign purposes. Potential solutions could be using a fake image detector, such as our curation network after pretraining or other works \cite{zhu2015learning,wang2019detecting,wang2020cnn} in order to determine whether the image has been edited, and if so where. 

In conclusion, we presented a benchmark and inpainting framework that allows us to dramatically outperform existing models at modern sensor resolutions. One key idea we had is to use a curation module to picks a good inpainted image by carefully comparing and contrasting visually similar images. The techniques used in our pipeline point to many different avenues of potential future work.  }

\input{supplemental}


\clearpage
%
%
\bibliographystyle{splncs04}
\bibliography{submission}


\end{document}

%% file: macros.tex
\def\IsArxiv{}      

\newcommand{\shortcite}[1]{\cite{#1}}

\ifdefined\IsArxiv
\newcommand{\arxiv} [1] {#1}
\newcommand{\eccvorarxiv} [2] {#2}
\else
\newcommand{\arxiv} [1] {}
\newcommand{\eccvorarxiv} [2] {#1}
\fi
\newcommand{\ignorethis} [1] {}

\newcommand{\sectnum    } [1] {\ref{#1}}

\newcommand{\sect       } [1] {Section~\sectnum{#1}}

\newcommand{\tbl}[1]{Table~\ref{#1}}

\newcommand{\fignum     } [1] {\ref{#1}}
\newcommand{\fig        } [1] {Figure~\fignum{#1}}

\definecolor{first}{rgb}{0,0.5,0}
\definecolor{second}{rgb}{0.6,0.4,0}
\definecolor{third}{rgb}{0.5,0,0}
\definecolor{egreen}{rgb}{0.1,0.6,0.5}
\definecolor{red}{rgb}{1.0,0.0,0.0}

\definecolor{verydarkgreen}{rgb}{0,0.2,0}
\definecolor{verydarkorange}{rgb}{0.4,0.2,0}
\definecolor{verydarkyellow}{rgb}{0.55,0.55,0}

\ifdefined\ShowNotes
\newcommand{\Kevin}[1]{\colornote{verydarkorange}{CB:}{#1}}
\newcommand{\Connelly}[1]{\colornote{darkgreen}{CB:}{#1}}
\newcommand{\Lingzhi}[1]{\colornote{dblue}{LZ}{#1}}
\newcommand{\Sohrab}[1]{\colornote{maroon}{SA}{#1}}
\newcommand{\es}[1]{\colornote{egreen}{ES}{#1}}
\newcommand{\zl}[1]{\colornote{darkred}{ZL}{#1}}
\newcommand{\todo}[1]{\colornote{verydarkgreen}{Todo}{#1}}
\else
\newcommand{\Kevin}[1]{}
\newcommand{\Connelly}[1]{}
\newcommand{\Lingzhi}[1]{}
\newcommand{\Sohrab}[1]{}
\newcommand{\es}[1]{}
\newcommand{\zl}[1]{}
\newcommand{\todo}[1]{}
\fi

\newcommand{\better}[1]{{ \color{darkgreen}\small($+$#1\%)}}
\newcommand{\worse}[1]{{ \color{darkred} \small($-$#1\%)}}

\ifdefined\ShowNotes
  \newcommand{\colornote}[3]{{\color{#1}\bf{#2: #3}\normalfont}}
\else
  \newcommand{\colornote}[3]{}
\fi

\definecolor{darkred}{rgb}{0.7,0.1,0.1}
\definecolor{darkgreen}{rgb}{0.1,0.5,0.1}
\definecolor{verydarkgreen}{rgb}{0.0,0.5,0.0}
\definecolor{verydarkblue}{rgb}{0.0,0.0,0.7}
\definecolor{cyan}{rgb}{0.7,0.0,0.7}
\definecolor{dblue}{rgb}{0.2,0.2,0.8}
\definecolor{maroon}{rgb}{0.76,.13,.28}
\definecolor{burntorange}{rgb}{0.81,.33,0}


%% file: figs/teaser.tex
\maketitle
\begin{center}
    \centering
    \includegraphics[trim=0.0in 2.8in 0.0in 0in, clip,width=4.8in]{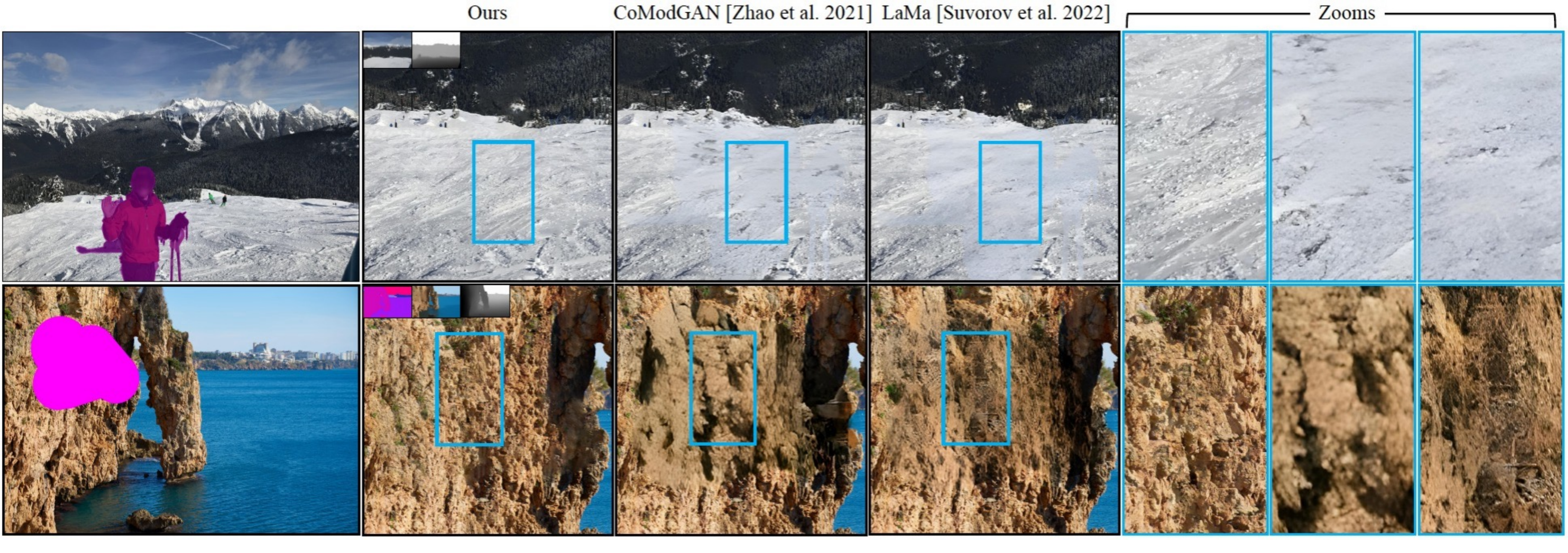}
    \vspace{-20 pt}
    \captionof{figure}{Inpainting at modern camera resolutions via guided PatchMatch (guides shown in inset) with a novel automatic curation module. Photos are 12 and 30 MP. Our result has 
    \emph{significantly} higher fidelity high-res detail than  CoModGAN~\cite{zhao2021comodgan} and LaMa~\cite{suvorov2021resolution}, the strongest baselines according to our user study, which were upsampled by Real-ESRGAN~\cite{wang2021real}. The guides on the top left of the second columns indicate the chosen guides for the specific image. }
    \label{fig:teaser}
\end{center}%

%% file: figs/overview.tex
\begin{figure*}[h]
    \centering
    \includegraphics[trim=0.0in 1.5in 0.0in 0.0in, clip,width=4.8in]{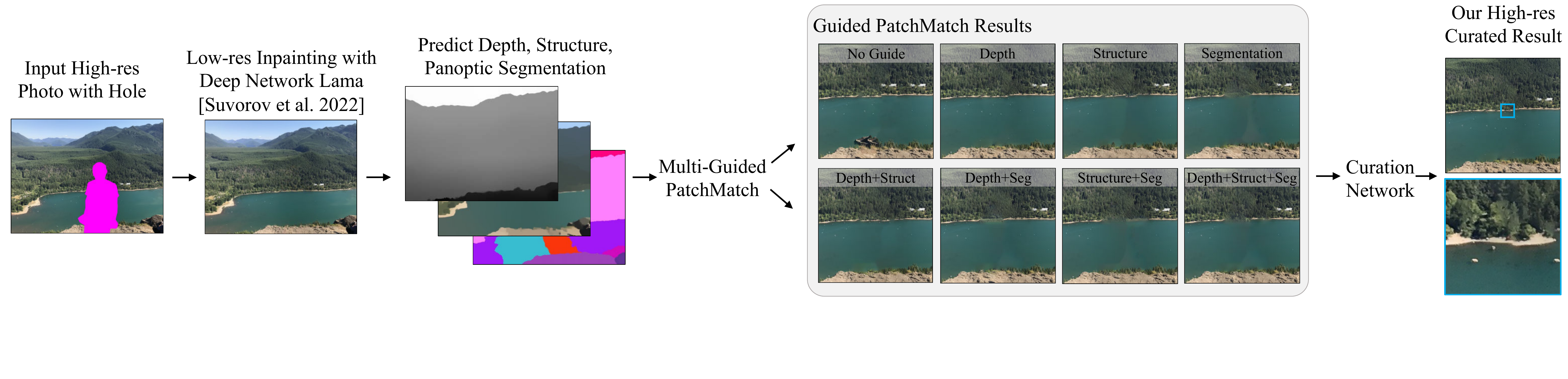}
    \caption{Overview of our framework. See the intro for discussion of components.}
    \label{fig:overview}
\end{figure*}


%% file: figs/curation_module.tex
\begin{figure*}
 \vspace{-15pt}
    \centering
    \includegraphics[trim=0.0in 0.0in 0.0in 0.0in, clip,width=4.0in]{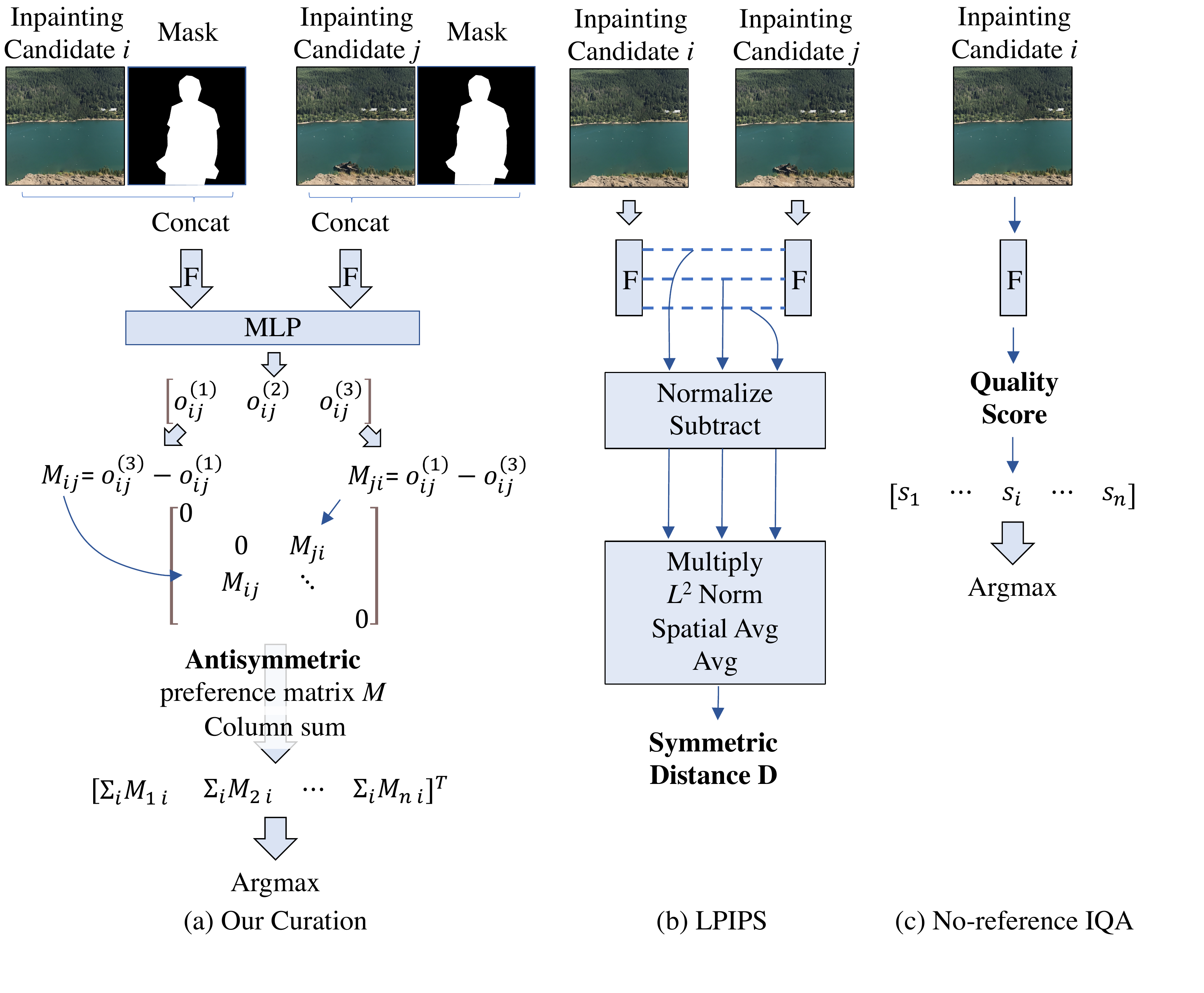}
    \vspace{-5ex}
    \caption{Our curation architecture produces scores in an \textbf{antisymmetric} matrix $M$ that is column-summed. LPIPS~\cite{Zhang_2018_CVPR} produces a \textbf{symmetric} distance score. Architectures used in RealismCNN~\cite{zhu2015learning} and no-reference IQA produce a score for each candidate image that is \textbf{independent} of the other candidates.}
    \label{fig:curation_module}
    \vspace{-17pt}
\end{figure*}

%% file: tables/curation_v2.tex
\setlength{\tabcolsep}{3pt}
\begin{table}
    \begin{center}
     \resizebox{0.8\textwidth}{!}{
      \begin{tabular}{l|l|l}
        \toprule 
        \textbf{Method} & \textbf{Accuracy} & \textbf{Accuracy for}\\
        & & \textbf{Easy Cases} \\
        \midrule 
        Human Performance & 57.1\% \better{0.7} & 86.1\% \better{1.4} \\
        \arrayrulecolor{black!30}\midrule
        \textbf{Our Curation Network} & \textbf{56.4\%} & \textbf{84.7\%} \\
        Ours No Pretraining & 53.9\% \worse{2.5} & 78.8\% \worse{5.9} \\
        \small Ours Fewer Augmentations & 53.8\% \worse{2.7} & 77.8\% \worse{6.9} \\
        Ours No Mask & 52.6\% \worse{3.8} & 72.2\% \worse{12.5} \\
        \small Ours Late Fusion Variant & 52.6\% \worse{3.9} & 82.1\% \worse{2.6} \\
        Ours Early Fusion & 51.4\% \worse{5.1} & 77.8\% \worse{6.9} \\
        Ours Freeze Backbone & 43.7\% \worse{12.7} & 53.0\% \worse{33.1} \\
        NIMA\cite{talebi2018nima} w/ MLP & 41.4\% \worse{15.0} & 50.0\% \worse{34.7} \\
        MetaIQA+\cite{zhu2021generalizable} w/ MLP & 41.1\% \worse{15.3} & 50.4\% \worse{34.3} \\
        Random Chance & 33.1\% \worse{23.3} & 30.6\% \worse{54.2} \\
        \arrayrulecolor{black}\bottomrule 
      \end{tabular}
      } 
      \vspace{5 pt}
      \caption{Curation network performance on human paired preference data, ablations, and comparisons. We report two test set accuracies: over the whole dataset, and over only easy cases. Our network is competitive with humans and outperforms all alternatives. Parenthetical numbers are relative to our network.}
      \label{tab:curation_v2}
    \end{center}
    \vspace{-20 pt}
\end{table}

%% file: tables/comp_with_sota.tex
\begin{table*}[h!]
  \vspace{-7pt}
    \begin{center}
    \resizebox{\textwidth}{!}{
      \begin{tabular}
      {l|c|c|c|c|c|c|c|c|c}
        \toprule 
        \textbf{Methods} & \textbf{LPIPS} $\downarrow$ & \multicolumn{2}{ c |}{\textbf{FID} $\downarrow$} & \multicolumn{2}{ c |}{\textbf{P-IDS} $\uparrow$} & \multicolumn{2}{ c |}{\textbf{U-IDS} $\uparrow$} & \textbf{User Pref. $\uparrow$} & \textbf{User Pref. $\uparrow$} \\
         &  & Full & Patch & Full & Patch & Full & Patch & Full Image & Boundary Patch \\
        \midrule 
        EdgeConnect \cite{nazeri2019edgeconnect} & 0.05017 & 35.06 & 41.05 & 0.04 & 4.56 & 0.00 & 0.55 & - & - \\
        Deepfillv2 \cite{yu2019free} & 0.05295 & 32.87 & 36.06 & 5.54 & 5.47 & 1.35 & 0.84 & - & - \\
        MEDFE \cite{liu2020rethinking} & 0.05170 & 33.97 & 60.87 & 0.48 & 2.23 & 0.00 & 0.26 & - & - \\
        HiFill \cite{yi2020contextual} & 0.05213 & 34.39 & \color{third}31.74 & 4.15 & 5.20 & 0.75 & 0.97 & - & - \\
        CoModGAN \cite{zhao2021comodgan} & 0.05099 & 24.81 & 32.08 & \color{third}14.72 & \color{second}7.01 & \color{third}4.47 & \color{second}1.51 & \color{second}28 & \color{third}17 \\
        MADF \cite{zhu2021image} & \color{third}0.04773 & \color{third}23.62 & 33.21 & 10.48 & 6.81 & 2.14 & \color{third}1.48 & 6 & 12 \\
        ProFill \cite{zeng2020high} & 0.04783 & 24.25 & \color{second}31.26 & 11.35 & \color{third}6.89 & 2.26 & 1.31 & 10 & 16 \\
        LaMa \cite{suvorov2021resolution} & \color{second}0.04588 & \color{second}19.20 & 35.95 & \color{second}17.24 & 6.86 & \color{second}5.62 & 1.38 & \color{second}28 & \color{second}22 \\
        \midrule
        SuperCAF (Ours) & \color{first}0.04156 & \color{first}18.74 & \color{first}15.63 & \color{first}22.46 & \color{first}19.77 & \color{first}10.70 & \color{first}10.22 & \color{first}128 & \color{first}133 \\
        \bottomrule 
      \end{tabular}
      }
      \vspace{5 pt}
      \caption{A comparison study with the state-of-the-art inpainting methods. The top 3 methods are colored: {\color{first}1}, {\color{second}2}, {\color{third}3}. }
      \label{tab:comp_with_sota}
    \end{center}
    \vspace{-30 pt}
\end{table*}


%% file: tables/improve_4_deep_and_pm_random_ours.tex
\begin{table}
    \begin{center}
      \begin{minipage}{.48\linewidth}

     \resizebox{1\textwidth}{!}{
      \begin{tabular}{l|c|c|c}
        \toprule 
        \textbf{Methods} & \textbf{LPIPS $\downarrow$} & \textbf{FID $\downarrow$} & \textbf{User Pref. $\uparrow$} \\
        \midrule 
        ProFill \cite{zeng2020high} + SR \cite{wang2021real} & 0.0478 & 24.2589 & 21 \\
        ProFill \cite{zeng2020high} + Ours & \textbf{0.0425} & \textbf{20.4362} & \textbf{179}  \\
        \midrule
        CoModGAN \cite{zhao2021comodgan} + SR \cite{wang2021real} & 0.0510 & 24.8189 & 36 \\
        CoModGAN \cite{zhao2021comodgan} + Ours & \textbf{0.0430} & \textbf{19.9224} & \textbf{164}  \\ 
        \midrule 
        MADF \cite{zhu2021image} + SR \cite{wang2021real} & 0.0477 & 23.6290 & 27 \\
        MADF \cite{zhu2021image} + Ours & \textbf{0.0421} & \textbf{19.9022} & \textbf{173}  \\
        \midrule
        LaMa \cite{suvorov2021resolution} + SR \cite{wang2021real} & 0.04588 & 19.2022 & 42 \\
        LaMa \cite{suvorov2021resolution} + Ours & \textbf{0.04156} & \textbf{18.7414} & \textbf{158} \\
        \bottomrule 
      \end{tabular}
      }
      \caption{A pairwise comparison study between high-resolution inpainting results for upsampled by Real-ESRGAN \cite{wang2021real} and results upsampled by our framework, for four recent inpainting methods. Users are asked to choose the best image from a pair for each inpainting model. The best score is bold.  }
      \label{tab:improve_4_deep_models}
      \end{minipage}\hspace{1em}%
      \begin{minipage}{.48\linewidth}
      \resizebox{1\textwidth}{!}{
      \begin{tabular}{l|c|c|c}
        \toprule 
        \textbf{Methods} & \textbf{LPIPS $\downarrow$} & \textbf{FID $\downarrow$} & \textbf{User Pref. $\uparrow$} \\
        \midrule 
        Content-Aware Fill & 0.04675 & 23.0068 & 16 \\
        Random Guided PM & 0.04281 & 19.7704 & 24 \\
        SuperCAF (Ours) & \textbf{0.04156} & \textbf{18.7414} & \textbf{60} \\
        \bottomrule 
      \end{tabular}
      }
      \caption{A comparison study with Photoshop's Content-Aware Fill, which is based on PatchMatch~\cite{barnes2009patchmatch} and a randomly selected Guided PatchMatch baseline. Photos average 20 megapixels and holes average 2.3 megapixels. The user studies are performed on 100 photographs where all method outputs are different from each other. The best scores are bold text in the table.}
      \label{tab:pm_random_ours}
      \end{minipage}
      
      
    \end{center}
    \vspace{-8ex}
\end{table}

%% file: figs/qualitative_results.tex
\begin{figure*}[h]
 \vspace{-10 pt}
    \centering
    \includegraphics[trim=0in 0in 0in 0in, clip,width=4.8in]{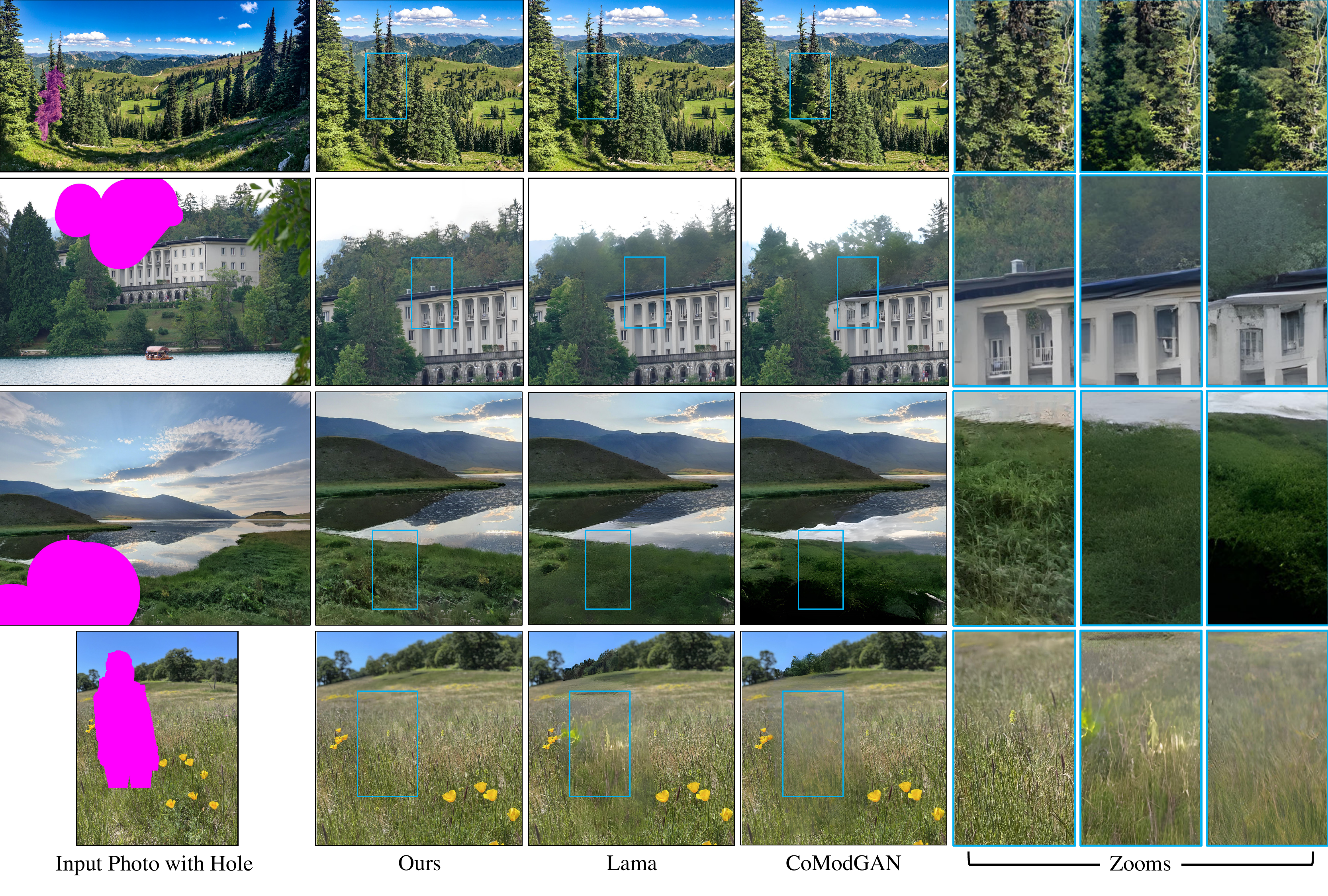}
    \vspace{-16 pt}
    \caption{Results for our method and two baselines with Real-ESRGAN, for (from top): a 26 MP nature panorama with real hole, a 20 MP photo with random scribble hole, a 12 MP lake photo with random scribble hole, a 12 MP field photo with an instance hole. Please check out supplemental for visual results.}
    \label{fig:qualitative_results}
    \vspace{-15 pt}
\end{figure*}

%% file: supplemental.tex
\title{Supplemental Materials: \\ Inpainting at Modern Camera Resolution by Guided PatchMatch with Auto-Curation} 

\titlerunning{ECCV-22 submission ID \ECCVSubNumber} 
\authorrunning{ECCV-22 submission ID \ECCVSubNumber} 
\author{Anonymous ECCV submission}
\institute{Paper ID \ECCVSubNumber}

\titlerunning{SuperCAF}
%
\author{Lingzhi Zhang\inst{1} \and Connelly Barnes\inst{2} \and Kevin Wampler\inst{2} \and Sohrab Amirghodsi\inst{2} \and \\ Eli Shechtman\inst{2} \and Zhe Lin\inst{2} \and Jianbo Shi\inst{1}}

\authorrunning{Zhang et al.}
%
\institute{University of Pennsylvania \and
Adobe Research \\
}
\maketitle

\section{Full Resolution Results and Comparisons}

Because of our research topic of inpainting at modern camera resolutions, we have prepared a 2 gigabyte supplemental material with an included HTML viewer that shows results of all methods at full resolution, and have uploaded that in an anonymized way to cloud file hosting. We checked with a program committee member and were told we can include a link as long as we are very confident we have anonymized it, which we are. The link is \url{https://drive.google.com/file/d/1Lmar1byASRReJ0SimBfWAE2Dt_-gMhor} . To prove that the supplemental is not changed since the time of submission, the MD5 sum of the 2 gigabyte supplemental material is 59eb1591b1601f491bd6962e43e81672 .

\section{Statistical Hypothesis Testing}

For the user preferences in Tables 2, 3, and 4 of the main paper, since users preferred our method, we tested whether this preference is statistically significant. We formed 10 null hypotheses: that ours in Table 2 was preferred equally to each of the four baselines, that ours in Table 3 was preferred equally to each of the four baselines, and that ours in Table 4 was preferred equally to the two baselines. We used a one sample permutation t test with $10^6$ simulations and found the $p$ values are all $0.0$ except in Table 4 the $p$ value for ours against Random-Guided PatchMatch is $p=0.000107$. After a Bonferroni correction the preference for our method is significant in all cases at a $p$ threshold of $0.01$. 

\section{Curation Network Details}

We generate a dataset for pretraining the curation network as follows: we first collect 48229 diverse photos taken by the authors and collaborators in many countries both indoors and outdoors, where the photos are 2K resolution or above, and resize them to 2K. For each image, we generate 10 synthetic holes using the same process and hole dataset described in ProFill~\cite{zeng2020high}: we generate 5 random stroke holes and 5 object-shaped holes, where each hole is constrained to fit in a randomly generated 512x512 bounding box. Then, for each of these holes, we compute all 8 possible guided PatchMatch results from our pipeline. In total, this results in more than 3 million inpainted images at 2K resolution. To reduce disk space requirements, we first generate the full inpainted 2K image in memory, and then crop it to the 512x512 hole bounding box before saving it to disk. We also collect the corresponding real photo crops before the synthetic holes were added. Additionally, we associate with each fake inpainted image the hole mask used to generate it, and for the corresponding real inpainted image with the same crop bounding box, we associate it with the hole mask used to produce the fake inpainting. We split the dataset into 80\% training, 10\% validation, and 10\% testing images.

When pretraining and when fine-tuning on human preferences, for data augmentations, we use horizontal and vertical flip, crop and resize, Gaussian blur, color jitter, Gaussian noise, rotation, and JPEG compression. When fine-tuning on human preferences, we use a cross-entropy classification loss and train until after the validation accuracy peaks and declines due to the model overfitting to the training set.

Some key lessons we learned in collecting our human preference data are that one should hire photographers either amateur or professional, train them well with meetings where one explains which of a pair of photos is better and why, and validate the quality of their preference data against a reference ground truth. In our case, the reference ground truth was established by an author who has a passion for photography. In previous variants of our data collection, we tried to use Amazon Mechanical Turk workers, but found that even when averaging opinions among many workers, the results were close to random chance. We also tried working one-on-one with expert human labelers who were not photographers, but we found their accuracy was much worse than if the workers were photographers.

\section{Automatic Cropping Details}

The automatic cropping mentioned in the main paper Section 3.4 works by placing a crop square around the hole and iteratively expanding the crop square from an initial size $s_i$ by a factor $\gamma=1.05$ until one of two conditions is hit: (1) the crop square is equal size or larger than the image along either dimension, (2) the number of hole pixels as determined by summation is less than a fraction $\tau$ of the crop square's pixels (we use $\tau=0.25$). The initial size is $s_i = \max(512, h_w, h_h)$, where $h_w, h_h$ are the width and height of the hole bounding box. Within each iteration, the crop square is first centered at the center of the hole bounding box, but in case the crop square moves outside the image, a translation is applied to each axis independently. This translation is such that for each axis, the crop square is moved back entirely inside the image with as few pixels of translation as possible, or if that is not possible, moves as few pixels as possible so all image pixels are visible in the crop along that axis.

\section{Discussion of Inpainting Evaluation Metrics}

It is well-known that image inpainting lacks good evaluation metrics. Previous inpainting works often use two types of quantitative evaluation metrics. The first type are the direct image content comparison metrics, such as LPIPS \cite{Zhang_2018_CVPR}, PSNR, and SSIM \cite{wang2004image}, which measure the similarity between pairs of inpainted images and original images. The other type are metrics that operate over the distributions of features within the dataset, such as FID \cite{parmar2021buggy} and P/U-IDS \cite{zhao2021comodgan}. FID measures distribution similarity on the deep features extracted from Inception network \cite{szegedy2016rethinking} for a set of inpainted images and a set of natural images. Similarly, P/U-IDS trains a linear SVM using both deep features of inpainted images and natural images, and classifies whether a deep Inception feature of an inpainted image is real or fake to show how realistic the inpainted images are. 

Unlike super resolution or image restoration tasks, we feel that direct image content comparison between the inpainted and original images does not truly reflect the inpainted image quality, especially when the holes are large. The reason is that the inpainted content could looks natural and realistic while simultaneously being very different from the original content. In practice, we indeed observe that in many cases that blurry and unnatural results have better scores for the content comparison metrics than the results that are natural but differ greatly from the original image. Clearly, this departs from human preference. Additionally, human do not really need to look at the original image as reference to tell whether an inpainted image is realistic, so this once again indicates that content comparison metrics might not be an appropriate metric for image inpainting. Thus, we think that metrics FID and P/U-IDS are relatively more appropriate and closer to human perception. This is consistent with the evidence presented in CoModGAN~\cite{zhao2021comodgan} that FID correlates highly with human preference rate as does P/U-IDS. Nevertheless, we report all metrics for the comparison with other state-of-the-art methods in section 4.4, and also conducted extensive user studies in the main paper. We believe that designing better quantitative metrics that tailor to inpainting task could be very useful for the community in the future works. 

\section{How Do Different Guides Influence Results?}

We show an example of how different guides might influence results in Figure \ref{fig:8_guided_outputs}. In this example, the depth guide (also chosen by our curation module) helps PatchMatch find the reference patches at the similar depth  and defocus to synthesize consistent texture. In general, different guides may be particularly useful in different cases, as we discuss in detail in the main paper.

\input{figs/8_guided_outputs}

\section{Additional Quantitative and User Study Experiments}

As an extension of Table 3 in the paper, we show that our method can also effectively boost the performance with older deep inpainting models, as shown in \tbl{tab:improve_3_deep_models}. Although we encourage inpainting models to be evaluated on benchmarks such as ours that correspond to modern camera sensors, we also show in \tbl{tab:comp_with_sota_different_res} results on our same benchmark dataset for lower resolutions 1K and 2K, which correspond to camera sensors released approximately \emph{two-and-a-half to two decades ago}, respectively~\cite{obsolete1,obsolete2}. We still find our method is always preferred the most according to the user studies at 1K and 2K, always preferred by the quantitative metrics at 2K, and is usually in first or second place according to quantitative metrics at 1K. Thus, our method helps the most at modern sensor resolutions but also behaves in a graceful way as resolution is lowered. 

\input{tables/improve_3_deep_models}

\input{tables/compare_with_sota_different_res}

\section{Additional Ablation Studies}

\subsection{Inpainting on Cropped Images vs. Full Images}

Here we study whether it will be helpful if deep inpainting models are run on cropped patches centered around the hole rather than on the full images. The cropped patches are generated by using the auto crop mechanism discussed previously. As shown in \tbl{tab:full_vs_crop}, we observe that the results generated from full image inpainting slightly outperforms the cropped variant. We believe that the main reason for this is that the holes used in our experiments are sufficiently large, which also leads to fairly large cropped patch. Thus, inpaintings on the cropped patches will need to go through a similar amount of subsequent upsampling as inpainting on the full images, however, full images have relatively more context. Due to our observations in this study, we report the inpainting results from deep inpainting models running on the full images in the main manuscript.

\input{tables/full_vs_crop}

\subsection{LaMa Inference on 2K Images}

The concurrent work LaMa \cite{suvorov2021resolution} mentions that their model can run directly on higher-resolution 2K images than the 256x256 images that it was trained on. We found that for our experiments, the best high-resolution inpainting results are obtained by running Lama \cite{suvorov2021resolution} by resizing down to a maximum size of 512 $\times$ 512 while preserving aspect ratio combined with a larger scale upsampling by Real-ESRGAN \cite{wang2021real}. In \tbl{tab:lama_2K_vs_512SR}, we show a high-resolution quantitative comparison between LaMa \cite{suvorov2021resolution} computing 2K output with a 2$\times$ SR and LaMa \cite{suvorov2021resolution} computing 512 $\times$ 512 output with a 8$\times$ SR. The results show that LaMa \cite{suvorov2021resolution} computed on 512 $\times$ 512 with a 8$\times$ SR outperforms the alternative on most metrics. A typical visual comparison between the two options can be found in \fig{fig:lama_2k_vs_512}. In general, we observe that LaMa \cite{suvorov2021resolution} computed at 512 resolution generates both better texture and more coherent structure reconstruction than the alternative. Therefore, based on our quantitative and qualitative investigations, we report the results from LaMa \cite{suvorov2021resolution} computing at 512 resolution in the main manuscript.

\input{tables/lama_2K_vs_512SR}

\input{figs/lama_2k_vs_512}

\subsection{Comparison with Bicubic Upsampling}

The most naive way to upsample the inpainted outputs from deep network is bicubic upsampling. In this section, for the LaMa \cite{suvorov2021resolution} model, we compare bicubic upsampling with Real-ESRGAN \cite{suvorov2021resolution} both qualitatively and quantitatively. The visual results in \fig{fig:bicubic_vs_realsr} shows a trade-off that the bicubic upsampling approach generates very blurry results compared to Real-ESRGAN \cite{suvorov2021resolution} outputs, while Real-ESRGAN \cite{suvorov2021resolution} sometimes produces slightly sharper boundary around the hole. Quantitatively, \tbl{tab:comp_bicubic_upsample} shows that Real-ESRGAN \cite{suvorov2021resolution} outperforms bicubic upsampling. Thus, we chose Real-ESRGAN \cite{suvorov2021resolution} over bicubic upsampling for upsampling the competing methods.

\input{tables/comp_bicubic_upsample}

\input{figs/bicubic_vs_realsr}

\subsection{Reference-Based Super Resolution}

To upsample the outputs of competing methods, we chose Real-ESRGAN \cite{suvorov2021resolution} since it is the state-of-the-arts super resolution algorithm and is also robust to visual artifacts in the natural image inputs. We also wondered whether reference-based super resolution could do a better job at upsampling the inpainting outputs of baselines. To answer this question, we tried a state-of-the-arts reference-based super resolution method named TTSR \cite{yang2020learning}. In the reference-based SR setting, we feed inpainted images with a maximum size of 512 $\times$ 512 (maintaining aspect ratio) as inputs and the 2K original image with hole as the reference images, generate the 4$\times$ upsampled outputs at 2K resolution, and finally upsample the outputs 2$\times$ to 4K resolution with Real-ESRGAN \cite{wang2021real}. Both methods share the same low resolution inpainted images computed from LaMa \cite{suvorov2021resolution}.

\input{tables/comp_ref_sr}

\input{figs/refsr_vs_realsr}

The current reference-based super resolution tasks mainly focus on upsampling images from like 128 to 512, while our starting resolution is 512. When we try to upsample images from 512 to 2048 with TTSR model \cite{yang2020learning}, it requires approximately 118 GB per image, which cannot fit in available GPU memory. This leaves us the only option to run the inference code on CPU. Due to the extremely slow computation time, we did this “TTSR vs. Real-ESRGAN” comparison study on a subset of 650 randomly sampled test images, which cost around 3 days computation time on a AMD Ryzen Threadripper 3960X 24-Core CPU. As shown in \tbl{tab:comp_ref_sr}, the quantitative results show that Real-ESRGAN \cite{wang2021real} is better than TTSR \cite{yang2020learning} in terms of FID and P/U-IDS and slightly worse in terms of LPIPS. Qualitatively, we observe that TTSR tends to consistently generate bumpy artifacts that are visually more unnatural than the Real-ESRGAN \cite{wang2021real} outputs, as shown in \fig{fig:refsr_vs_realsr}. Therefore, based on the quality comparison and computation feasibility, we use Real-ESRGAN \cite{wang2021real} as the upsampling method for the competing methods in our paper.

\subsection{Guided PatchMatch using RGB Images}

While our proposed method uses depth, structure map, and panoptic segmentation computed as the guides for PatchMatch, we also evaluate the inpainting quality when using the RGB image produced by the deep inpainting backbone as the guide directly. As shown in \fig{fig:rgb_vs_ours}, we observe that the RGB guided PatchMatch tend to produce very blurry results, and thus make the texture visually unnatural. The quantitative evaluation shown in \tbl{tab:rgb_guide} also indicates that the RGB guided PatchMatch outputs do not achieve very good results. Thus, we decided not to incorporate the RGB guided PatchMatch as an option in our curation stage. More discussion of guide choices can be found in section 3.2 in the main manuscript.

\input{tables/RGB_guide}

\input{figs/rgb_vs_ours}

\subsection{All Metrics}

As promised in the main manuscript, we show all the evaluation metrics for the comparison with competing methods in \tbl{tab:comp_with_sota_supp}. More discussion of the results can be found in Section 4.3 in the main manuscript, and discussion of evaluation metrics can be found in Section 3 in the supplemental material.

\input{tables/comp_with_sota}
\section{Failure Cases}

As discussed in the last section of our main manuscript, our method has some limitations. In this section, we provide several visual demonstration in \fig{fig:failure_cases} to show the limitations we mentioned in the main paper, and more discussion can be found in the caption of \fig{fig:failure_cases}. 

\input{figs/failure_cases}

\section{More Qualitative Results}

We show more inpainting results at 4K or above resolution in \fig{fig:4K_inpaint_1}, \fig{fig:4K_inpaint_2}, \fig{fig:4K_inpaint_3}, and \fig{fig:4K_inpaint_4} in the following pages.

\input{figs/4K_inpaint_1}
\input{figs/4K_inpaint_2}
\input{figs/4K_inpaint_3}
\input{figs/4K_inpaint_4}




%% file: figs/8_guided_outputs.tex
\begin{figure}[!h]
    \centering
    \includegraphics[trim=0.0in 0.0in 0.0in 0.0in, clip,width=4.8in]{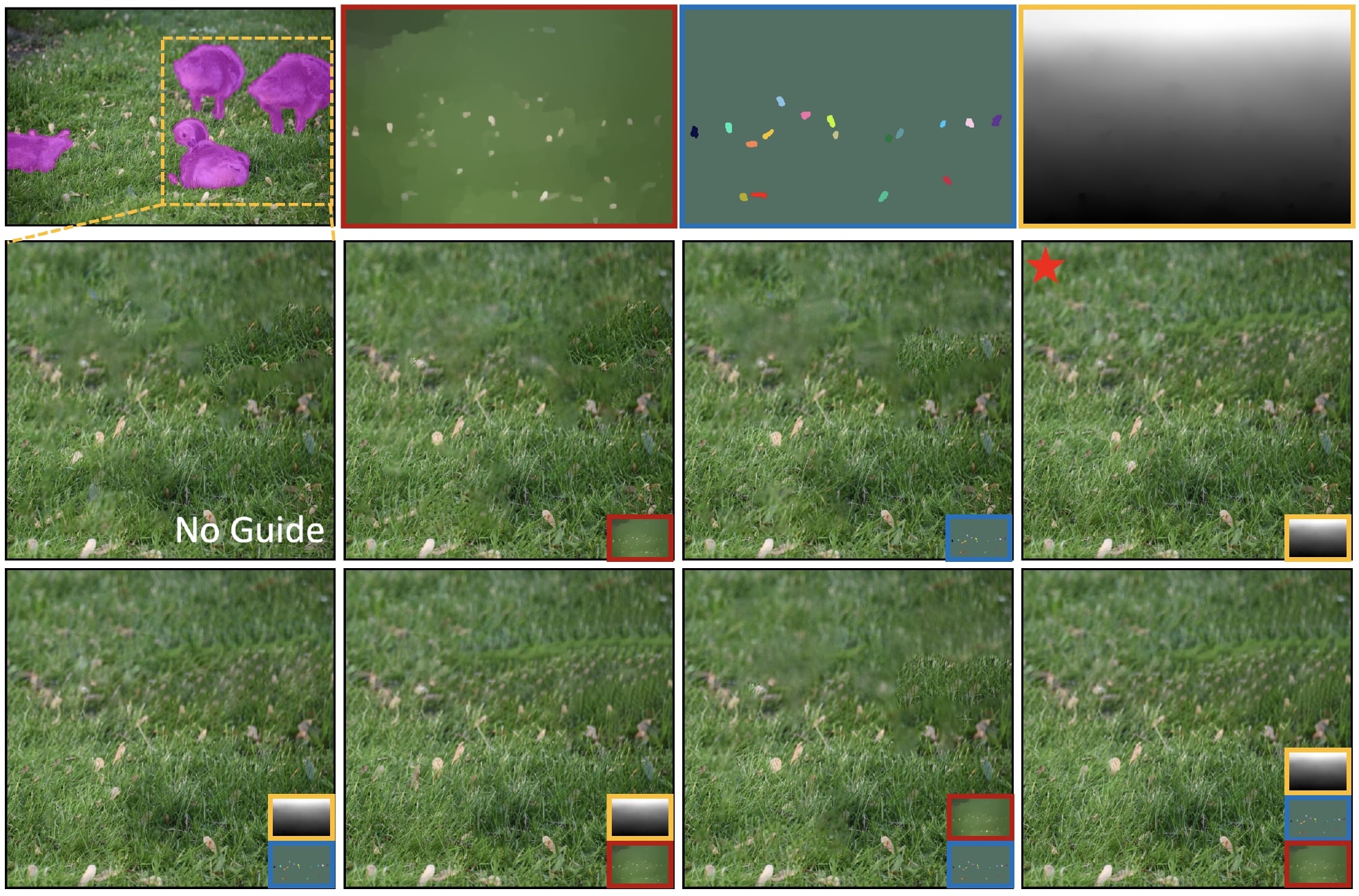}
    \caption{The top row shows the input image with hole and the guides, and the middle and bottom rows show the 8 guided PatchMatch results using different guides. \textbf{{\color{red}\Large{$\star$}}} represents the guided result that was chosen by the curation module. The results that use a depth guide generally avoid copying patches that are defocused or have incorrect texture scale. \textbf{Please zoom in to see the details.} }
    \label{fig:8_guided_outputs}
\end{figure}

%% file: tables/improve_3_deep_models.tex
\begin{table}[!h]
    \begin{center}
     \resizebox{0.6\textwidth}{!}{
      \begin{tabular}{l|c|c}
        \toprule 
        \textbf{Methods} & \textbf{LPIPS $\downarrow$} & \textbf{FID $\downarrow$} \\
        \midrule 
        MEDFE + SR / \textbf{Ours} & 0.05170 / \textbf{0.04442} & 33.97 / \textbf{21.81} \\ 
        \midrule 
        EdgeConnect  + SR  / \textbf{Ours} & 0.05017 / \textbf{0.04350} & 35.06 / \textbf{21.77}   \\
        \midrule
        Deepfillv2 + SR / \textbf{Ours} & 0.05295 / \textbf{0.04349} & 32.87 / \textbf{20.50}  \\
        \bottomrule 
      \end{tabular}
      }
      \caption{A comparison for older models between results upsampled by Real-ESRGAN and results upsampled by our proposed method.}
      \label{tab:improve_3_deep_models}
    \end{center}
\end{table}

%% file: tables/compare_with_sota_different_res.tex
\begin{table}[!h]
    \begin{center}
     \resizebox{1\textwidth}{!}{
      \begin{tabular}
      {l|c|c|c|c|c|c|c|c}
        \toprule 
        \textbf{Methods} & \textbf{LPIPS} $\downarrow$ & \multicolumn{2}{ c |}{\textbf{FID} $\downarrow$} & \multicolumn{2}{ c |}{\textbf{P-IDS} $\uparrow$} & \multicolumn{2}{ c |}{\textbf{U-IDS} $\uparrow$} & {\textbf{User Pref.} $\uparrow$} \\
         &  & Full & Patch & Full & Patch & Full & Patch & Full \\
        \midrule 
        CoMoD & 0.0598/0.0617 & 20.76/23.10 & {\color{verydarkgreen}13.99}/{\color{verydarkgreen}19.26} & {\color{verydarkgreen}21.57}/{\color{verydarkgreen}18.88} & {\color{verydarkgreen}19.85}/{\color{verydarkgreen}13.89} & {\color{verydarkgreen}10.10}/{\color{verydarkgreen}8.70} & {\color{verydarkgreen}7.55}/{\color{verydarkgreen}4.54} & {\color{verydarkgreen}22}/{\color{verydarkgreen}21} \\
        MADF & {\color{verydarkgreen}0.0548}/{\color{verydarkgreen}0.0564} & {\color{verydarkgreen}20.02}/{\color{verydarkgreen}22.24} & 15.21/21.76 & 15.96/12.85 & 14.33/10.75 & 5.05/3.53 & 3.92/2.60 & 13/10 \\
        ProFill & 0.0631/0.0574 & 20.64/22.83 & 16.52/22.37 & 16.37/13.97 & 12.80/10.59 & 5.75/3.94 & 2.86/2.66 & {\color{verydarkgreen}22}/14 \\
        LaMa & {\color{darkred}0.0509}/{\color{verydarkblue}0.0537} & {\color{verydarkblue}17.10}/{\color{verydarkblue}18.43} & {\color{verydarkblue}11.82}/{\color{verydarkblue}20.05} & {\color{darkred}23.77}/{\color{verydarkblue}20.72} & {\color{verydarkblue}20.73}/{\color{verydarkblue}13.96} & {\color{darkred}12.04}/{\color{verydarkblue}9.33} & {\color{verydarkblue}8.05}/{\color{verydarkblue}4.13} & {\color{verydarkblue}59}/{\color{verydarkblue}36} \\
        \midrule
        Ours & {\color{verydarkblue}0.0522}/{\color{darkred}0.0525} & {\color{darkred}16.01}/{\color{darkred}17.88} & {\color{darkred}10.23}/{\color{darkred}12.41} & {\color{verydarkblue}23.62}/{\color{darkred}22.88} & {\color{darkred}22.83}/{\color{darkred}20.82} & {\color{verydarkblue}11.95}/{\color{darkred}11.51} & {\color{darkred}9.78}/{\color{darkred}9.87} & {\color{darkred}84}/{\color{darkred}119} \\
        \bottomrule 
      \end{tabular}
       }
      \caption{A comparison study with the state-of-the-art methods at lower resolution of 1K / 2K corresponding to older camera sensors released approximately \emph{two-and-a-half and two decades ago}, respectively~\cite{obsolete1,obsolete2}. The top 3 methods are colored: {\color{darkred}1}, {\color{verydarkblue}2}, {\color{verydarkgreen}3}. }
      \label{tab:comp_with_sota_different_res}
    \end{center}
\end{table}

%% file: tables/full_vs_crop.tex
\begin{table}[h!]
    \begin{center}
     \resizebox{0.6\textwidth}{!}{
      \begin{tabular}{l|c|c|c|c}
        \toprule 
        \textbf{Methods} & \textbf{LPIPS $\downarrow$} & \textbf{FID $\downarrow$} & \textbf{P-IDS $\uparrow$} & \textbf{U-IDS $\uparrow$} \\
        \midrule 
        CoModGAN \cite{zhao2021comodgan} on Crop & 0.05244 & 25.05 & \textbf{16.68} & \textbf{5.46} \\
        CoModGAN \cite{zhao2021comodgan} on Full & \textbf{0.05099} & \textbf{24.8} & 14.72 & 4.47 \\
        \midrule 
        MADF \cite{zhu2021image} on Crop & 0.04774 & 28.21 & 4.52 & 0.50 \\
        MADF \cite{zhu2021image} on Full & \textbf{0.04773} & \textbf{23.62} & \textbf{10.48} & \textbf{2.14} \\
        \midrule 
        ProFill \cite{zeng2020high} on Crop & 0.04925 & 24.34 & 10.32 & 2.72 \\
        ProFill \cite{zeng2020high} on Full & \textbf{0.04783} & \textbf{24.25} & \textbf{11.35} & \textbf{2.26} \\
        \midrule 
        LaMa \cite{suvorov2021resolution} on Crop & 0.04600 & \textbf{19.19} & 17.03 & 5.43 \\
        LaMa \cite{suvorov2021resolution} on Full & \textbf{0.04588} & 19.20 & \textbf{17.24} & \textbf{5.62} \\
        \bottomrule 
      \end{tabular}
      }
      \caption{Quantitative comparison between inpainting methods running on cropped patches vs. full images. }
      \label{tab:full_vs_crop}
    \end{center}
\end{table}

%% file: tables/lama_2K_vs_512SR.tex

\begin{table}[h!]
    \begin{center}
     \resizebox{0.6\textwidth}{!}{
      \begin{tabular}{l|c|c|c|c}
        \toprule 
        \textbf{Methods} & \textbf{LPIPS $\downarrow$} & \textbf{FID $\downarrow$} & \textbf{P-IDS $\uparrow$} & \textbf{U-IDS $\uparrow$} \\
        \midrule 
        LaMa \cite{suvorov2021resolution} on 2K & \textbf{0.03982} & 25.53 & 5.18 & 1.18 \\
        LaMa \cite{suvorov2021resolution} on 512 & 0.04588 & \textbf{19.20} & \textbf{17.24} & \textbf{5.62}  \\
        \bottomrule 
      \end{tabular}
      }
      \caption{Comparison between LaMa \cite{suvorov2021resolution} direct computing on 2K resolution images with 2$\times$ super resolution and LaMa \cite{suvorov2021resolution} computing on 512 images with 8$\times$ super resolution. Both cases use the Real-ESRGAN \cite{wang2021real} for super resolution step. }
      \label{tab:lama_2K_vs_512SR}
    \end{center}
    \vspace{-15 pt}
\end{table}

%% file: figs/lama_2k_vs_512.tex
\begin{figure}[!h]
    \centering
    \includegraphics[trim=3.0in 2.0in 3.0in 0in, clip,width=4.8in]{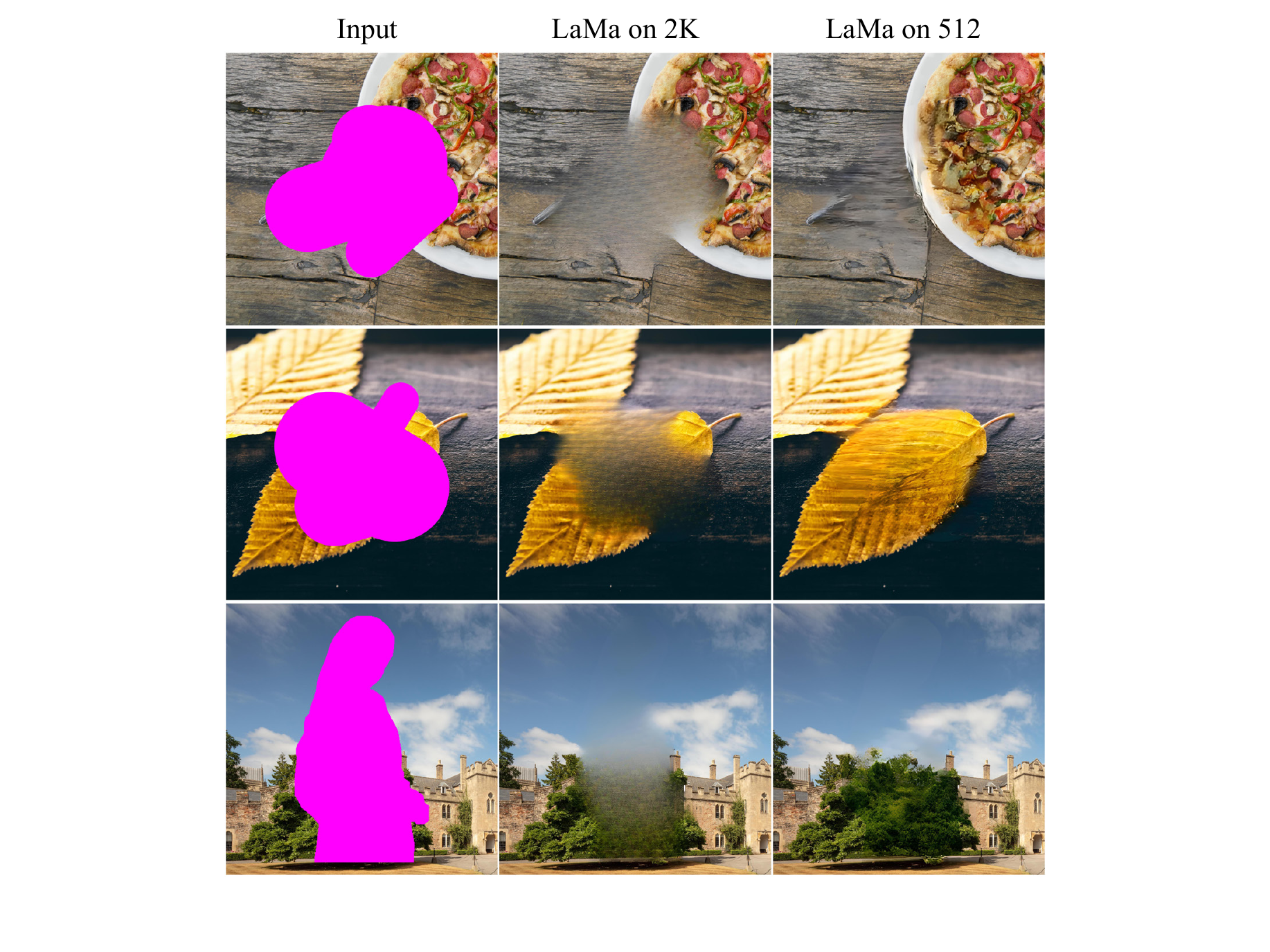}
    \vspace{-30 pt}
    \caption{Visual comparison between LaMa \cite{suvorov2021resolution} run on 2K images with 2$\times$ SR and LaMa \cite{suvorov2021resolution} run on maximum size of 512 $\times$ 512 images with 8$\times$ SR. }
    \label{fig:lama_2k_vs_512}
\end{figure}

%% file: tables/comp_bicubic_upsample.tex
\begin{table}[h!]
    \begin{center}
     \resizebox{0.6\textwidth}{!}{
      \begin{tabular}{l|c|c|c|c}
        \toprule 
        \textbf{Methods} & \textbf{LPIPS $\downarrow$} & \textbf{FID $\downarrow$} & \textbf{P-IDS $\uparrow$} & \textbf{U-IDS $\uparrow$} \\
        \midrule 
        + Bicubic Upsampling & 0.05043 & 23.66 & 16.46 & 5.14 \\
        + Real-ESRGAN \cite{wang2021real} & \textbf{0.04588} & \textbf{19.20} & \textbf{17.24} & \textbf{5.62} \\
        \bottomrule 
      \end{tabular}
      }
      \caption{Quantitative comparison between upsampled inpainting results using bicubic upsampling and Real-ESRGAN \cite{wang2021real}. }
      \label{tab:comp_bicubic_upsample}
    \end{center}
\end{table}

%% file: figs/bicubic_vs_realsr.tex
\begin{figure}[!h]
    \centering
    \includegraphics[trim=3.0in 2.0in 3.0in 0in, clip,width=4.8in]{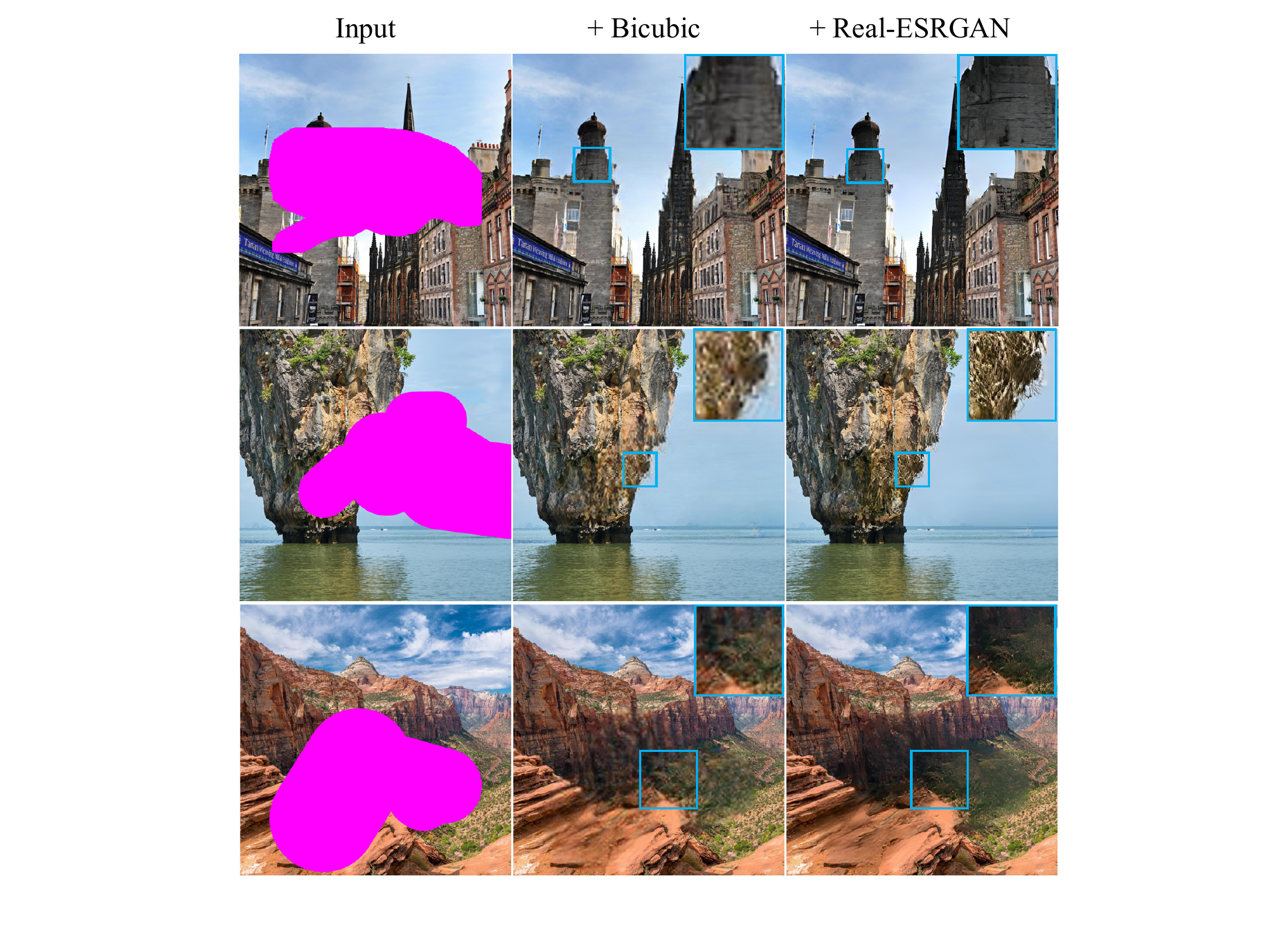}
    \vspace{-30 pt}
    \caption{Qualitative comparison for the LaMa \cite{suvorov2021resolution} model between upsampled inpainting results using bicubic upsampling and Real-ESRGAN \cite{wang2021real}.}
    \label{fig:bicubic_vs_realsr}
\end{figure}

%% file: tables/comp_ref_sr.tex
\begin{table}[h!]
    \begin{center}
     \resizebox{0.6\textwidth}{!}{
      \begin{tabular}{l|c|c|c|c}
        \toprule 
        \textbf{Methods} & \textbf{LPIPS $\downarrow$} & \textbf{FID $\downarrow$} & \textbf{P-IDS $\uparrow$} & \textbf{U-IDS $\uparrow$} \\
        \midrule 
        + TTSR \cite{yang2020learning} & \textbf{0.04391} & 25.93 & 3.09 & 1.03 \\
        + Real-ESRGAN \cite{wang2021real} & 0.04571 & \textbf{25.03} & \textbf{6.33} & \textbf{1.66} \\
        \bottomrule 
      \end{tabular}
      }
      \caption{Quantitative comparison between upsampled inpainting results using TTSR \cite{wang2021real} and Real-ESRGAN \cite{wang2021real}. }
      \label{tab:comp_ref_sr}
    \end{center}
\end{table}

%% file: figs/refsr_vs_realsr.tex
\begin{figure}[!h]
    \centering
    \includegraphics[trim=3.0in 2.0in 3.0in 0in, clip,width=4.8in]{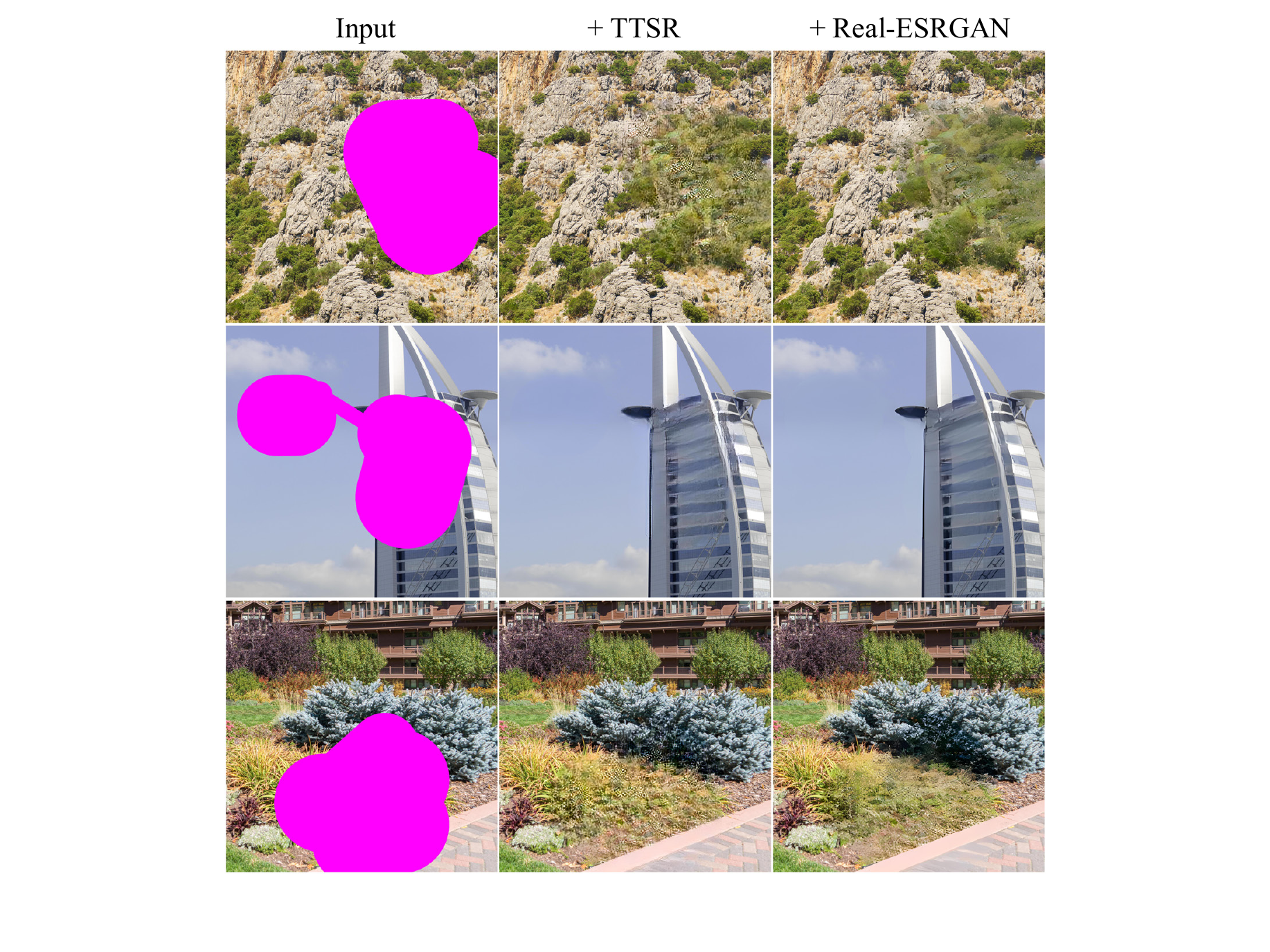}
    \vspace{-30 pt}
    \caption{Qualitative comparison between upsampled inpainting results using TTSR \cite{wang2021real} and Real-ESRGAN \cite{wang2021real}.}
    \label{fig:refsr_vs_realsr}
\end{figure}

%% file: tables/RGB_guide.tex
\begin{table}[h!]
    \begin{center}
     \resizebox{0.6\textwidth}{!}{
      \begin{tabular}{l|c|c|c|c}
        \toprule 
        \textbf{Methods} & \textbf{LPIPS $\downarrow$} & \textbf{FID $\downarrow$} & \textbf{P-IDS $\uparrow$} & \textbf{U-IDS $\uparrow$} \\
        \midrule 
        RGB Guide & \textbf{0.04112} & 24.65 & 16.39 & 5.41 \\
        Our Full Model & 0.04156 & \textbf{18.74} & \textbf{22.46} & \textbf{10.70} \\
        \bottomrule 
      \end{tabular}
      }
      \caption{Quantitative comparison between using RGB guided PatchMatch and our full model. }
      \label{tab:rgb_guide}
    \end{center}
     \vspace{-30 pt}
\end{table}

%% file: figs/rgb_vs_ours.tex
\begin{figure}[!h]
    \centering
    \includegraphics[trim=3.0in 2.0in 3.0in 0in, clip,width=4.8in]{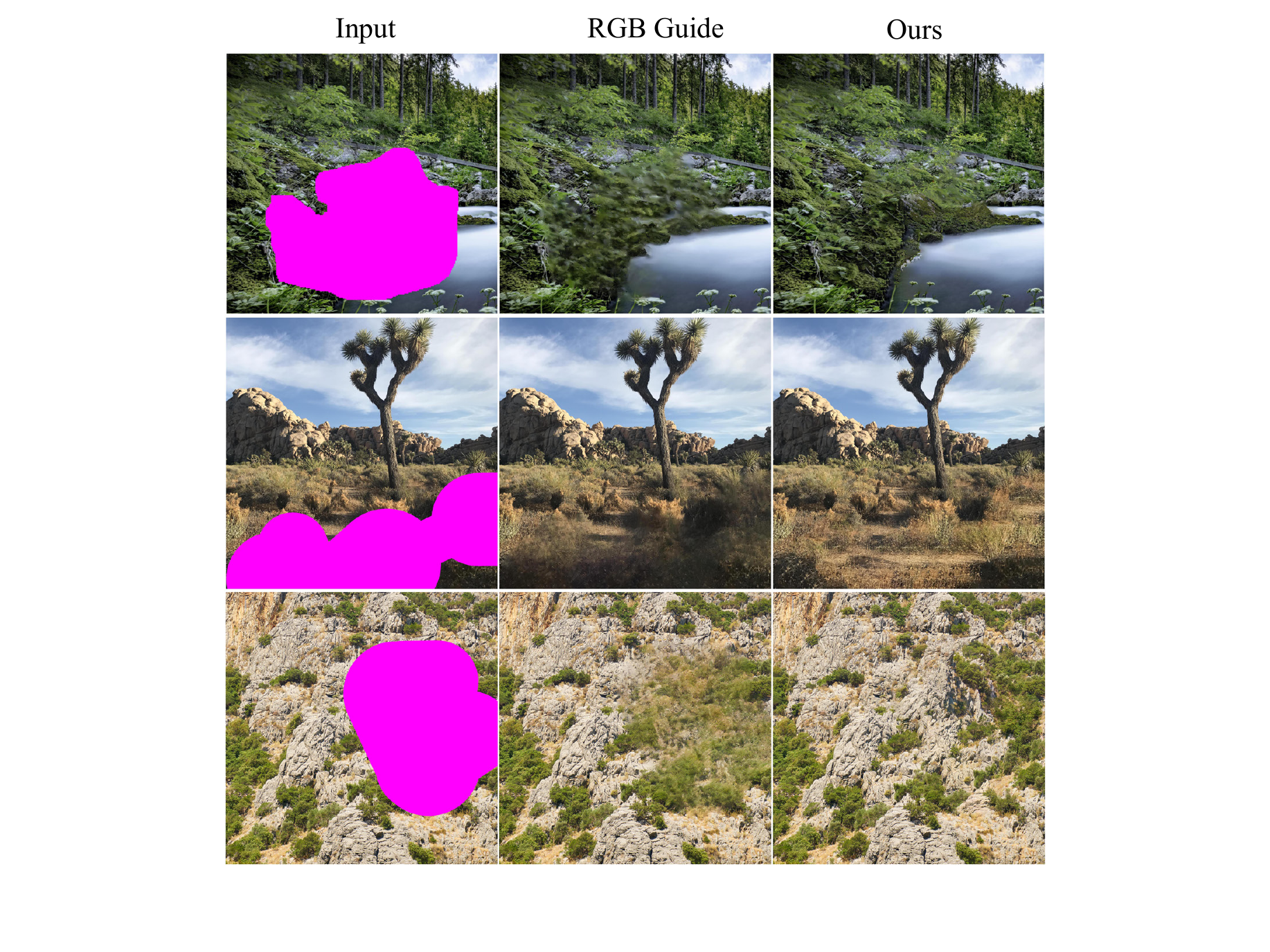}
    \vspace{-30 pt}
    \caption{Qualitative comparison between using RGB guided PatchMatch and our full model. }
    \label{fig:rgb_vs_ours}
\end{figure}

%% file: figs/failure_cases.tex
\begin{figure*}[!t]
    \centering
    \includegraphics[trim=0.2in 0in 0.2in 0in, clip,width=4.8in]{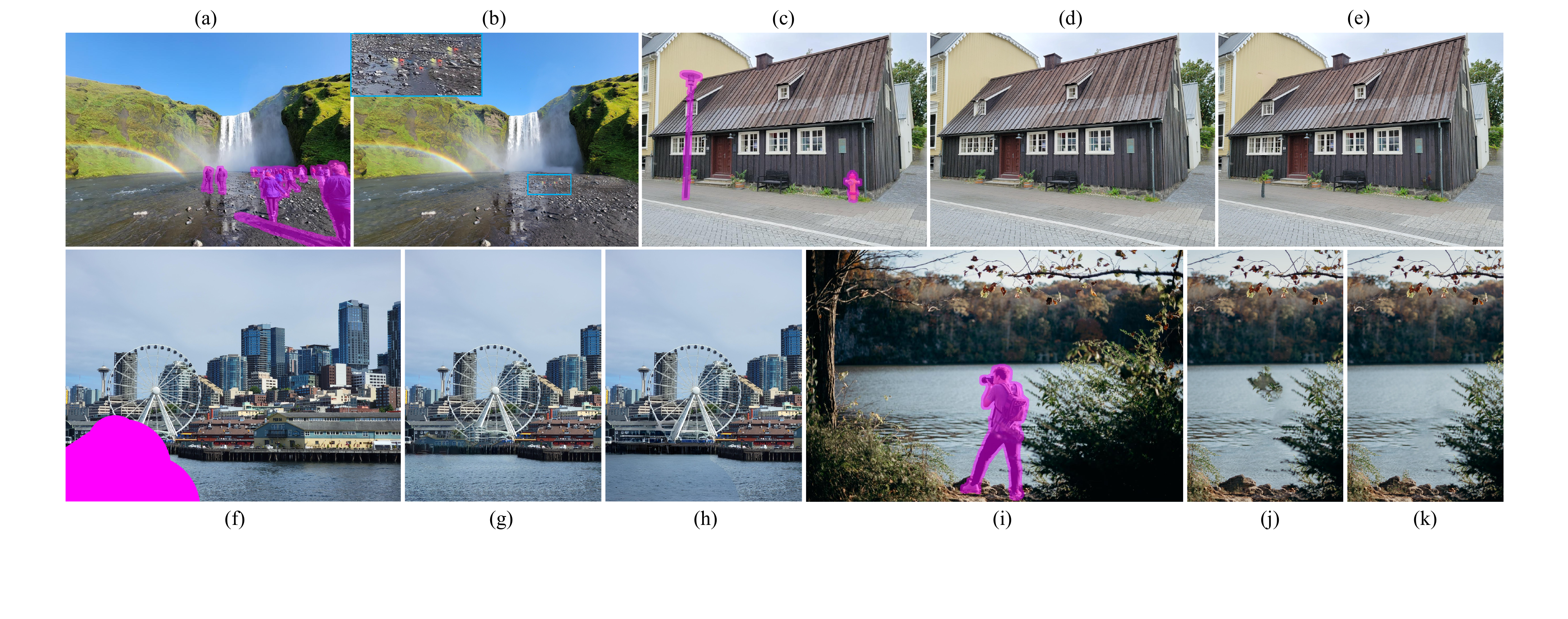}
    \vspace{-10 pt}
    \caption{Limitations of our method. In (a) and (b), undesirable repetitions of the salient yellow and red objects occur in the inpainted region. In (c) and (d), structures under perspective can be broken for the window at the top of the house, while LaMa \cite{suvorov2021resolution} (e) reconstructs better structure at the window region but has visual artifacts in other regions. In (f) and (g), our  patch-based synthesis method fails to inpaint the Ferris wheel, since the image lacks good reference content to directly copy from to fill the hole and requires hallucinating new structure. In this case, LaMa \cite{suvorov2021resolution} builds some new structure to better fill the Ferris wheel but produce over-smooth pixels in the water region, in (h). In (f) - (k), the curation network sometimes does not pick the best option from eight Guided PatchMatch outputs. In this case, the curation network picks (j) instead of (k), and thus a human manually picking  could help for this case. The last section of our main manuscript discusses ways these limitations could potentially be mitigated in future work.}
    \label{fig:failure_cases}
\end{figure*}

%% file: figs/4K_inpaint_1.tex
\begin{figure*}[!h]
    \centering
    \includegraphics[trim=0in 0.5in 0in 0in, clip,width=4.8in]{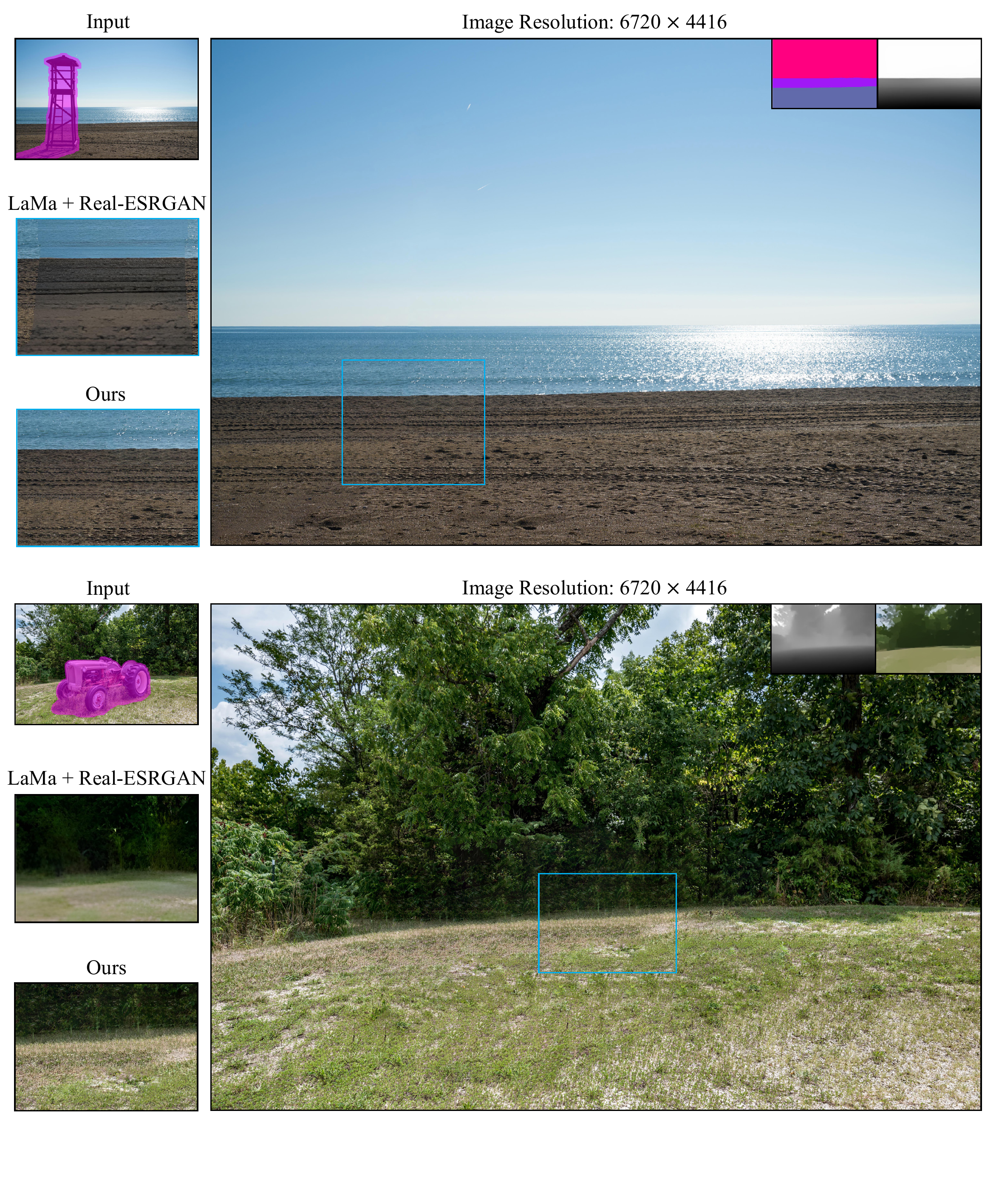}
    \vspace{-20 pt}
    \caption{In each of the two examples, the top left is the input image with a real object mask, the right giant image is the inpainted result from our method, the other two images on the bottom left are the zoom in insets for the closet competing method LaMa \cite{suvorov2021resolution} upsampled by Real-ESRGAN \cite{wang2021real} and ours, where the zoom in location are indicated by the bounding box on the right image. The guides used for each image is shown at the top right corner of the inpainted result.}
    \label{fig:4K_inpaint_1}
    \vspace{-10 pt}
\end{figure*}

%% file: figs/4K_inpaint_2.tex
\begin{figure*}[!h]
    \centering
    \includegraphics[trim=0in 0.5in 0in 0in, clip,width=4.8in]{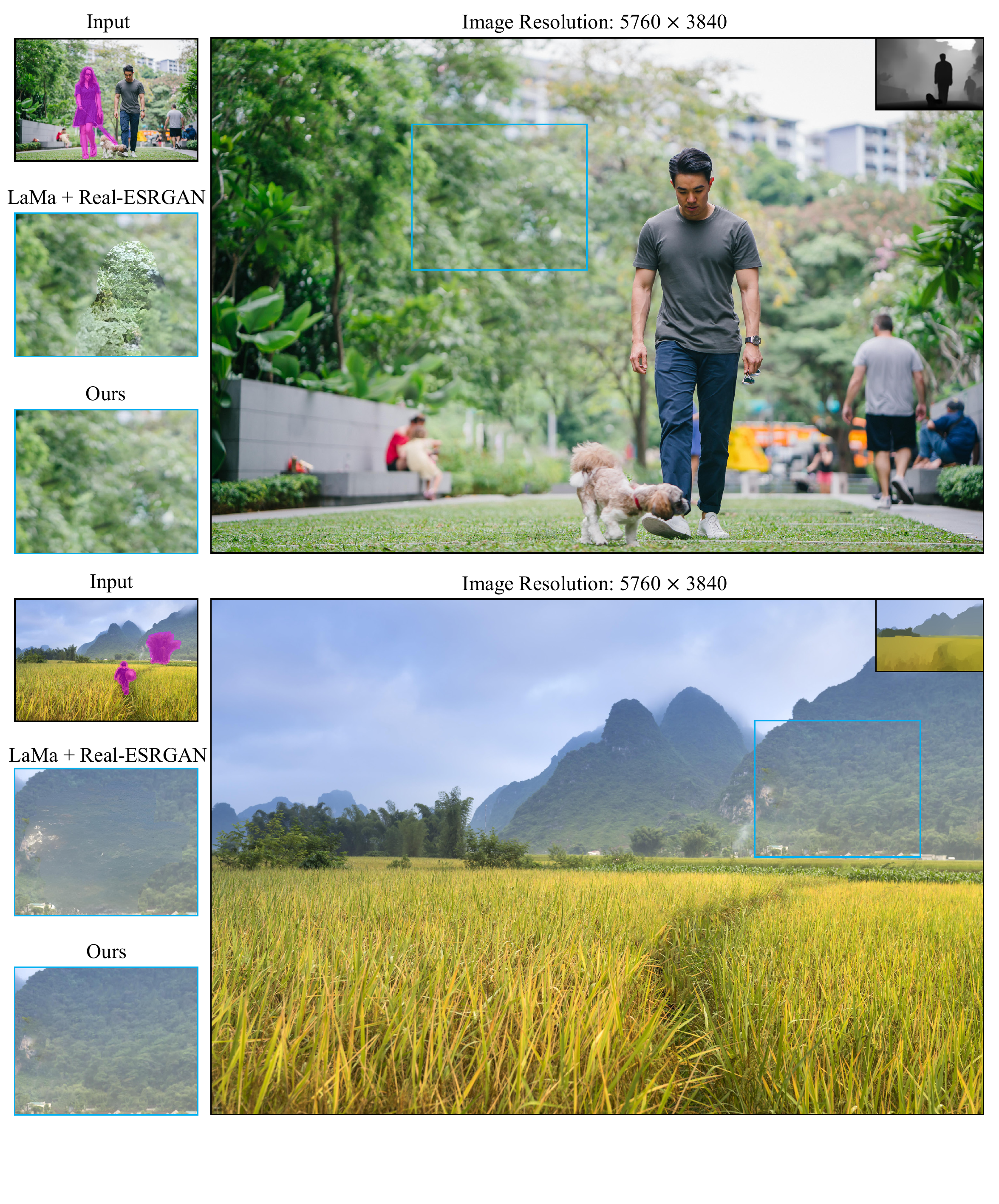}
    \vspace{-20 pt}
    \caption{In each of the two examples, the top left is the input image with a real object mask, the right giant image is the inpainted result from our method, the other two images on the bottom left are the zoom in insets for the closet competing method LaMa \cite{suvorov2021resolution} upsampled by Real-ESRGAN \cite{wang2021real} and ours, where the zoom in location are indicated by the bounding box on the right image. The guides used for each image is shown at the top right corner of the inpainted result.}
    \label{fig:4K_inpaint_2}
    \vspace{-10 pt}
\end{figure*}

%% file: figs/4K_inpaint_3.tex
\begin{figure*}[!h]
    \centering
    \includegraphics[trim=0in 0.5in 0in 0in, clip,width=4.8in]{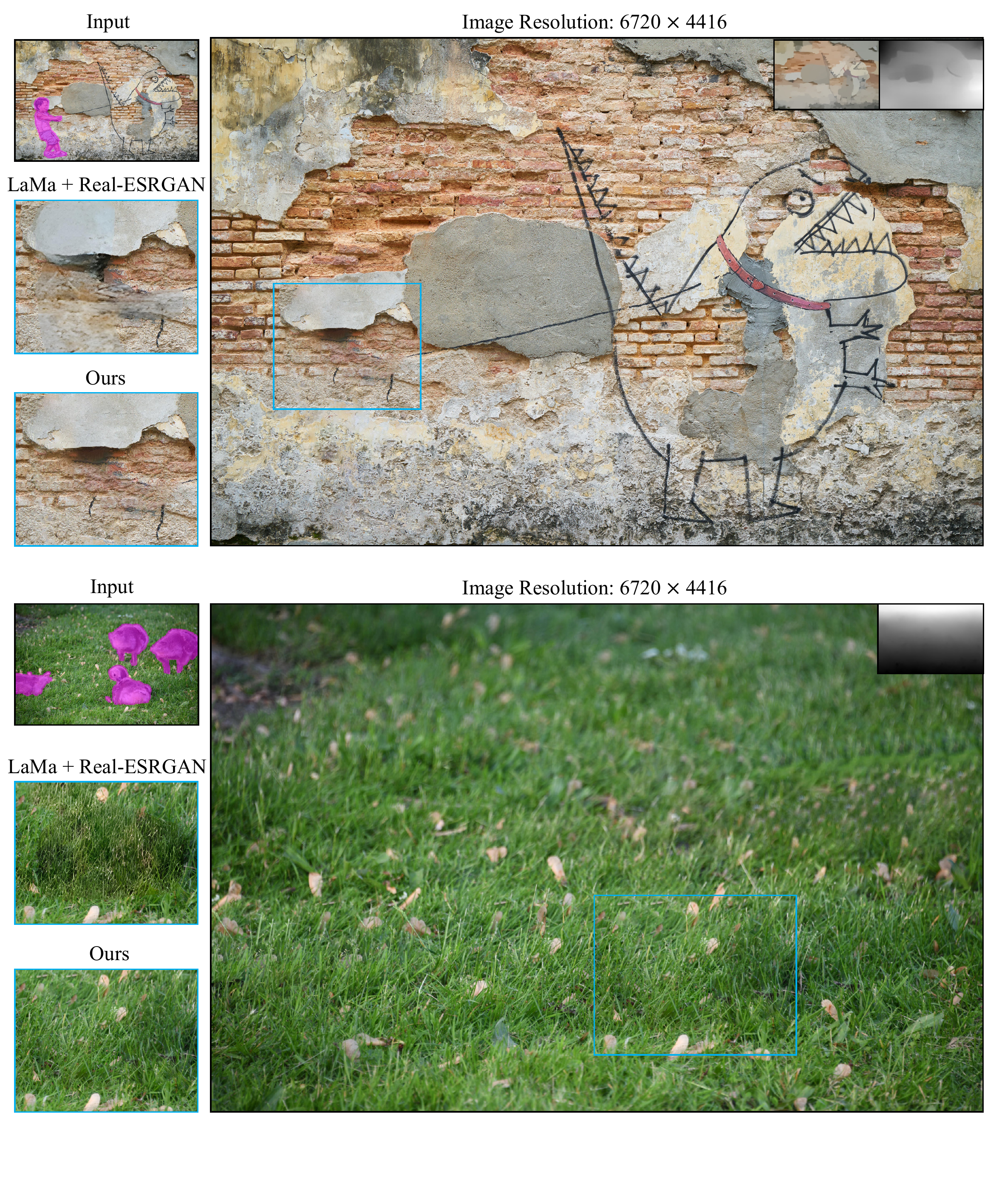}
    \vspace{-20 pt}
    \caption{In each of the two examples, the top left is the input image with a real object mask, the right giant image is the inpainted result from our method, the other two images on the bottom left are the zoom in insets for the closet competing method LaMa \cite{suvorov2021resolution} upsampled by Real-ESRGAN \cite{wang2021real} and ours, where the zoom in location are indicated by the bounding box on the right image. The guides used for each image is shown at the top right corner of the inpainted result.}
    \label{fig:4K_inpaint_3}
    \vspace{-10 pt}
\end{figure*}

%% file: figs/4K_inpaint_4.tex
\begin{figure*}[!h]
    \centering
    \includegraphics[trim=0in 0.5in 0in 0in, clip,width=4.8in]{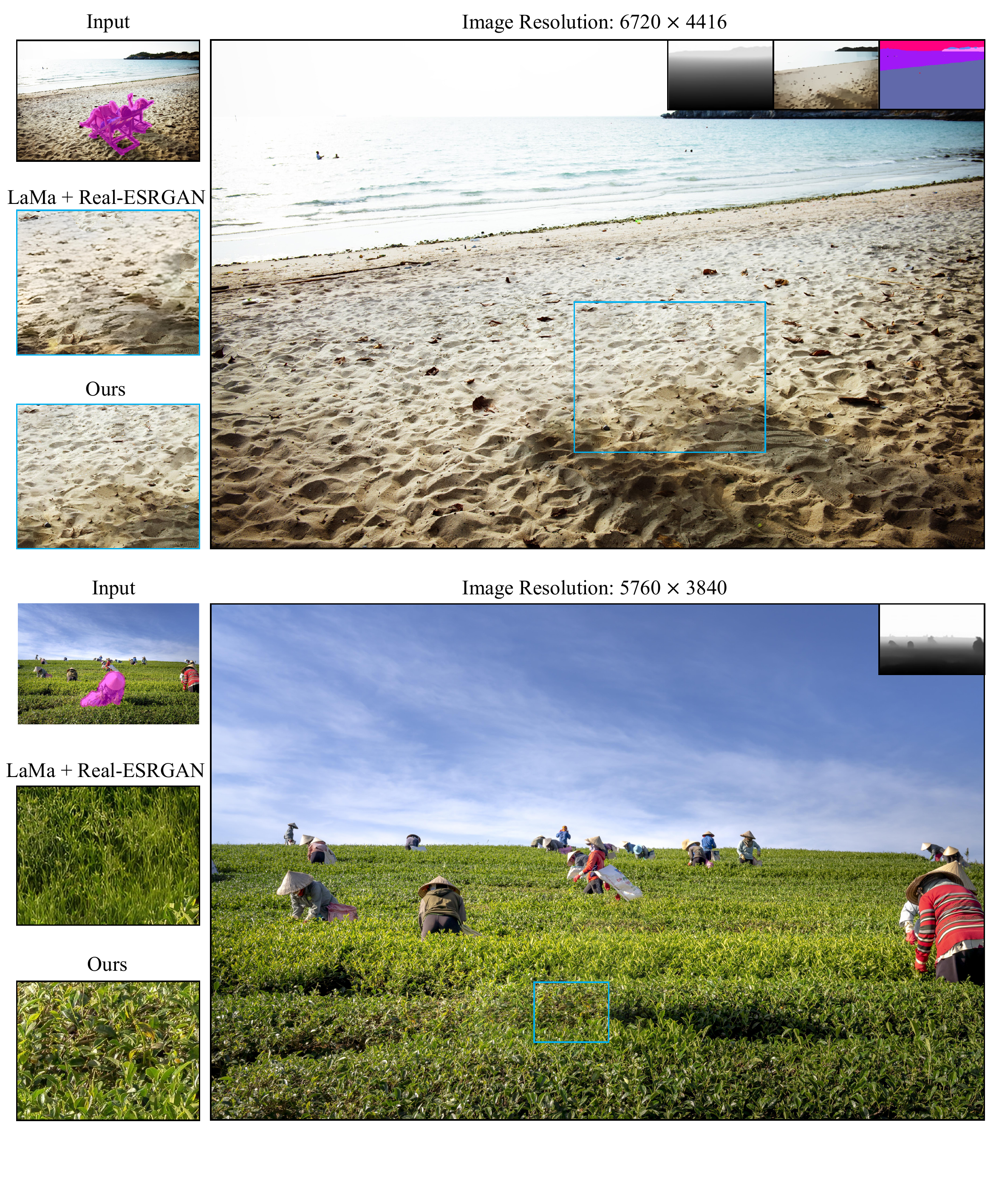}
    \vspace{-20 pt}
    \caption{In each of the two examples, the top left is the input image with a real object mask, the right giant image is the inpainted result from our method, the other two images on the bottom left are the zoom in insets for the closet competing method LaMa \cite{suvorov2021resolution} upsampled by Real-ESRGAN \cite{wang2021real} and ours, where the zoom in location are indicated by the bounding box on the right image. The guides used for each image is shown at the top right corner of the inpainted result.}
    \label{fig:4K_inpaint_4}
    \vspace{-10 pt}
\end{figure*}

%% file: submission.bbl
\begin{thebibliography}{10}
\providecommand{\url}[1]{\texttt{#1}}
\providecommand{\urlprefix}{URL }
\providecommand{\doi}[1]{https://doi.org/#1}

\bibitem{barnes2009patchmatch}
Barnes, C., Shechtman, E., Finkelstein, A., Goldman, D.B.: Patchmatch: A
  randomized correspondence algorithm for structural image editing. ACM Trans.
  Graph.  \textbf{28}(3), ~24 (2009)

\bibitem{benard2013stylizing}
B{\'e}nard, P., Cole, F., Kass, M., Mordatch, I., Hegarty, J., Senn, M.S.,
  Fleischer, K., Pesare, D., Breeden, K.: Stylizing animation by example. ACM
  Transactions on Graphics (TOG)  \textbf{32}(4),  1--12 (2013)

\bibitem{bosse2017deep}
Bosse, S., Maniry, D., M{\"u}ller, K.R., Wiegand, T., Samek, W.: Deep neural
  networks for no-reference and full-reference image quality assessment. IEEE
  Transactions on image processing  \textbf{27}(1),  206--219 (2017)

\bibitem{burt1987laplacian}
Burt, P.J., Adelson, E.H.: The laplacian pyramid as a compact image code. In:
  Readings in computer vision, pp. 671--679. Elsevier (1987)

\bibitem{fujix}
Cade, D.: The world’s first ‘fully’ digital camera was created by fuji
  (2016),
  \url{https://petapixel.com/2016/06/09/photo-history-worlds-first-fully-digital-camera-invented-fuji/}

\bibitem{obsolete2}
Canon: Canon camera museum: Powershot a75 (2022),
  \url{https://global.canon/en/c-museum/product/dcc495.html}

\bibitem{darabi2012image}
Darabi, S., Shechtman, E., Barnes, C., Goldman, D.B., Sen, P.: Image melding:
  Combining inconsistent images using patch-based synthesis. ACM Transactions
  on graphics (TOG)  \textbf{31}(4),  1--10 (2012)

\bibitem{diamanti2015synthesis}
Diamanti, O., Barnes, C., Paris, S., Shechtman, E., Sorkine-Hornung, O.:
  Synthesis of complex image appearance from limited exemplars. ACM
  Transactions on Graphics (TOG)  \textbf{34}(2),  1--14 (2015)

\bibitem{duggal2019deeppruner}
Duggal, S., Wang, S., Ma, W.C., Hu, R., Urtasun, R.: Deeppruner: Learning
  efficient stereo matching via differentiable patchmatch. In: Proceedings of
  the IEEE/CVF International Conference on Computer Vision. pp. 4384--4393
  (2019)

\bibitem{fivser2016stylit}
Fi{\v{s}}er, J., Jamri{\v{s}}ka, O., Luk{\'a}{\v{c}}, M., Shechtman, E.,
  Asente, P., Lu, J., S{\`y}kora, D.: Stylit: illumination-guided example-based
  stylization of 3d renderings. ACM Transactions on Graphics (TOG)
  \textbf{35}(4),  1--11 (2016)

\bibitem{gu2019div8k}
Gu, S., Lugmayr, A., Danelljan, M., Fritsche, M., Lamour, J., Timofte, R.:
  Div8k: Diverse 8k resolution image dataset. In: 2019 IEEE/CVF International
  Conference on Computer Vision Workshop (ICCVW). pp. 3512--3516. IEEE (2019)

\bibitem{obsolete1}
van Hall, D.: Digital kameramuseum: Canon powershot 600n (1997) (2022),
  \url{https://www.digitalkameramuseum.de/en/cameras/item/canon-powershot-600n}

\bibitem{he2012statistics}
He, K., Sun, J.: Statistics of patch offsets for image completion. In: European
  conference on computer vision. pp. 16--29. Springer (2012)

\bibitem{hertzmann2001image}
Hertzmann, A., Jacobs, C.E., Oliver, N., Curless, B., Salesin, D.H.: Image
  analogies. In: Proceedings of the 28th annual conference on Computer graphics
  and interactive techniques. pp. 327--340 (2001)

\bibitem{huang2014image}
Huang, J.B., Kang, S.B., Ahuja, N., Kopf, J.: Image completion using planar
  structure guidance. ACM Transactions on graphics (TOG)  \textbf{33}(4),
  1--10 (2014)

\bibitem{iizuka2017globally}
Iizuka, S., Simo-Serra, E., Ishikawa, H.: Globally and locally consistent image
  completion. ACM Transactions on Graphics (ToG)  \textbf{36}(4),  1--14 (2017)

\bibitem{jamrivska2019stylizing}
Jamri{\v{s}}ka, O., Sochorov{\'a}, {\v{S}}., Texler, O., Luk{\'a}{\v{c}}, M.,
  Fi{\v{s}}er, J., Lu, J., Shechtman, E., S{\`y}kora, D.: Stylizing video by
  example. ACM Transactions on Graphics (TOG)  \textbf{38}(4),  1--11 (2019)

\bibitem{kaspar2015self}
Kaspar, A., Neubert, B., Lischinski, D., Pauly, M., Kopf, J.: Self tuning
  texture optimization. In: Computer Graphics Forum. vol.~34, pp. 349--359.
  Wiley Online Library (2015)

\bibitem{li2021fully}
Li, Y., Zhao, H., Qi, X., Wang, L., Li, Z., Sun, J., Jia, J.: Fully
  convolutional networks for panoptic segmentation. In: Proceedings of the
  IEEE/CVF Conference on Computer Vision and Pattern Recognition. pp. 214--223
  (2021)

\bibitem{liao2017visual}
Liao, J., Yao, Y., Yuan, L., Hua, G., Kang, S.B.: Visual attribute transfer
  through deep image analogy. ACM Trans. Graph.  \textbf{36}(4),  120:1--120:15
  (Jul 2017). \doi{10.1145/3072959.3073683},
  \url{http://doi.acm.org/10.1145/3072959.3073683}

\bibitem{liu2018image}
Liu, G., Reda, F.A., Shih, K.J., Wang, T.C., Tao, A., Catanzaro, B.: Image
  inpainting for irregular holes using partial convolutions. In: Proceedings of
  the European Conference on Computer Vision (ECCV). pp. 85--100 (2018)

\bibitem{liu2020rethinking}
Liu, H., Jiang, B., Song, Y., Huang, W., Yang, C.: Rethinking image inpainting
  via a mutual encoder-decoder with feature equalizations. arXiv preprint
  arXiv:2007.06929  (2020)

\bibitem{liu2020real}
Liu, W., Zhang, P., Huang, X., Yang, J., Shen, C., Reid, I.: Real-time image
  smoothing via iterative least squares. ACM Transactions on Graphics (TOG)
  \textbf{39}(3),  1--24 (2020)

\bibitem{nazeri2019edgeconnect}
Nazeri, K., Ng, E., Joseph, T., Qureshi, F.Z., Ebrahimi, M.: Edgeconnect:
  Generative image inpainting with adversarial edge learning. arXiv preprint
  arXiv:1901.00212  (2019)

\bibitem{parmar2021buggy}
Parmar, G., Zhang, R., Zhu, J.Y.: On buggy resizing libraries and surprising
  subtleties in fid calculation. arXiv preprint arXiv:2104.11222  (2021)

\bibitem{pathak2016context}
Pathak, D., Krahenbuhl, P., Donahue, J., Darrell, T., Efros, A.A.: Context
  encoders: Feature learning by inpainting. In: Proceedings of the IEEE
  conference on computer vision and pattern recognition. pp. 2536--2544 (2016)

\bibitem{ranftl2021vision}
Ranftl, R., Bochkovskiy, A., Koltun, V.: Vision transformers for dense
  prediction. In: Proceedings of the IEEE/CVF International Conference on
  Computer Vision. pp. 12179--12188 (2021)

\bibitem{ren2019structureflow}
Ren, Y., Yu, X., Zhang, R., Li, T.H., Liu, S., Li, G.: Structureflow: Image
  inpainting via structure-aware appearance flow. In: Proceedings of the
  IEEE/CVF International Conference on Computer Vision. pp. 181--190 (2019)

\bibitem{sun2019high}
Sun, K., Zhao, Y., Jiang, B., Cheng, T., Xiao, B., Liu, D., Mu, Y., Wang, X.,
  Liu, W., Wang, J.: High-resolution representations for labeling pixels and
  regions. arXiv preprint arXiv:1904.04514  (2019)

\bibitem{suvorov2021resolution}
Suvorov, R., Logacheva, E., Mashikhin, A., Remizova, A., Ashukha, A.,
  Silvestrov, A., Kong, N., Goka, H., Park, K., Lempitsky, V.:
  Resolution-robust large mask inpainting with fourier convolutions. WACV:
  Winter Conference on Applications of Computer Vision  (2022)

\bibitem{szegedy2016rethinking}
Szegedy, C., Vanhoucke, V., Ioffe, S., Shlens, J., Wojna, Z.: Rethinking the
  inception architecture for computer vision. In: Proceedings of the IEEE
  conference on computer vision and pattern recognition. pp. 2818--2826 (2016)

\bibitem{talebi2018nima}
Talebi, H., Milanfar, P.: Nima: Neural image assessment. IEEE Transactions on
  Image Processing  \textbf{27}(8),  3998--4011 (2018)

\bibitem{tan2019efficientnet}
Tan, M., Le, Q.: Efficientnet: Rethinking model scaling for convolutional
  neural networks. In: International Conference on Machine Learning. pp.
  6105--6114. PMLR (2019)

\bibitem{wang2019detecting}
Wang, S.Y., Wang, O., Owens, A., Zhang, R., Efros, A.A.: Detecting photoshopped
  faces by scripting photoshop. In: Proceedings of the IEEE/CVF International
  Conference on Computer Vision. pp. 10072--10081 (2019)

\bibitem{wang2020cnn}
Wang, S.Y., Wang, O., Zhang, R., Owens, A., Efros, A.A.: Cnn-generated images
  are surprisingly easy to spot... for now. In: Proceedings of the IEEE/CVF
  Conference on Computer Vision and Pattern Recognition. pp. 8695--8704 (2020)

\bibitem{wang2021real}
Wang, X., Xie, L., Dong, C., Shan, Y.: Real-esrgan: Training real-world blind
  super-resolution with pure synthetic data. In: Proceedings of the IEEE/CVF
  International Conference on Computer Vision. pp. 1905--1914 (2021)

\bibitem{wang2004image}
Wang, Z., Bovik, A.C., Sheikh, H.R., Simoncelli, E.P.: Image quality
  assessment: from error visibility to structural similarity. IEEE transactions
  on image processing  \textbf{13}(4),  600--612 (2004)

\bibitem{wexler2007space}
Wexler, Y., Shechtman, E., Irani, M.: Space-time completion of video. IEEE
  Transactions on pattern analysis and machine intelligence  \textbf{29}(3),
  463--476 (2007)

\bibitem{xiong2019foreground}
Xiong, W., Yu, J., Lin, Z., Yang, J., Lu, X., Barnes, C., Luo, J.:
  Foreground-aware image inpainting. In: Proceedings of the IEEE/CVF Conference
  on Computer Vision and Pattern Recognition. pp. 5840--5848 (2019)

\bibitem{xu2012structure}
Xu, L., Yan, Q., Xia, Y., Jia, J.: Structure extraction from texture via
  relative total variation. ACM transactions on graphics (TOG)  \textbf{31}(6),
   1--10 (2012)

\bibitem{yang2020learning}
Yang, F., Yang, H., Fu, J., Lu, H., Guo, B.: Learning texture transformer
  network for image super-resolution. In: Proceedings of the IEEE/CVF
  Conference on Computer Vision and Pattern Recognition. pp. 5791--5800 (2020)

\bibitem{yi2020contextual}
Yi, Z., Tang, Q., Azizi, S., Jang, D., Xu, Z.: Contextual residual aggregation
  for ultra high-resolution image inpainting. In: Proceedings of the IEEE/CVF
  Conference on Computer Vision and Pattern Recognition. pp. 7508--7517 (2020)

\bibitem{yin2021CVPR}
Yin, W., Zhang, J., Wang, O., Niklaus, S., Mai, L., Chen, S., Shen, C.:
  Learning to recover 3d scene shape from a single image. In: Proc. IEEE Conf.
  Comp. Vis. Patt. Recogn. (CVPR) (2021)

\bibitem{yu2018generative}
Yu, J., Lin, Z., Yang, J., Shen, X., Lu, X., Huang, T.S.: Generative image
  inpainting with contextual attention. In: Proceedings of the IEEE conference
  on computer vision and pattern recognition. pp. 5505--5514 (2018)

\bibitem{yu2019free}
Yu, J., Lin, Z., Yang, J., Shen, X., Lu, X., Huang, T.S.: Free-form image
  inpainting with gated convolution. In: Proceedings of the IEEE/CVF
  International Conference on Computer Vision. pp. 4471--4480 (2019)

\bibitem{zeng2020high}
Zeng, Y., Lin, Z., Yang, J., Zhang, J., Shechtman, E., Lu, H.: High-resolution
  image inpainting with iterative confidence feedback and guided upsampling.
  In: European Conference on Computer Vision. pp. 1--17. Springer (2020)

\bibitem{zhang2020resnest}
Zhang, H., Wu, C., Zhang, Z., Zhu, Y., Lin, H., Zhang, Z., Sun, Y., He, T.,
  Mueller, J., Manmatha, R., et~al.: Resnest: Split-attention networks. arXiv
  preprint arXiv:2004.08955  (2020)

\bibitem{Zhang_2018_CVPR}
Zhang, R., Isola, P., Efros, A.A., Shechtman, E., Wang, O.: The unreasonable
  effectiveness of deep features as a perceptual metric. In: Proceedings of the
  IEEE Conference on Computer Vision and Pattern Recognition (CVPR) (June 2018)

\bibitem{zhang2018unreasonable}
Zhang, R., Isola, P., Efros, A.A., Shechtman, E., Wang, O.: The unreasonable
  effectiveness of deep features as a perceptual metric. In: Proceedings of the
  IEEE conference on computer vision and pattern recognition. pp. 586--595
  (2018)

\bibitem{zhao2021comodgan}
Zhao, S., Cui, J., Sheng, Y., Dong, Y., Liang, X., Chang, E.I., Xu, Y.: Large
  scale image completion via co-modulated generative adversarial networks. In:
  International Conference on Learning Representations (ICLR) (2021)

\bibitem{zhou2017places}
Zhou, B., Lapedriza, A., Khosla, A., Oliva, A., Torralba, A.: Places: A 10
  million image database for scene recognition. IEEE Transactions on Pattern
  Analysis and Machine Intelligence  (2017)

\bibitem{zhou2017scene}
Zhou, B., Zhao, H., Puig, X., Fidler, S., Barriuso, A., Torralba, A.: Scene
  parsing through ade20k dataset. In: Proceedings of the IEEE conference on
  computer vision and pattern recognition. pp. 633--641 (2017)

\bibitem{zhou2021cocosnet}
Zhou, X., Zhang, B., Zhang, T., Zhang, P., Bao, J., Chen, D., Zhang, Z., Wen,
  F.: Cocosnet v2: Full-resolution correspondence learning for image
  translation. In: Proceedings of the IEEE/CVF Conference on Computer Vision
  and Pattern Recognition. pp. 11465--11475 (2021)

\bibitem{zhou2021full}
Zhou, X., Zhang, B., Zhang, T., Zhang, P., Bao, J., Chen, D., Zhang, Z., Wen,
  F.: Full-resolution correspondence learning for image translation. Proc. IEEE
  Conf. Comp. Vis. Patt. Recogn. (CVPR)  (2021)

\bibitem{zhu2020metaiqa}
Zhu, H., Li, L., Wu, J., Dong, W., Shi, G.: Metaiqa: Deep meta-learning for
  no-reference image quality assessment. In: Proceedings of the IEEE/CVF
  Conference on Computer Vision and Pattern Recognition. pp. 14143--14152
  (2020)

\bibitem{zhu2021generalizable}
Zhu, H., Li, L., Wu, J., Dong, W., Shi, G.: Generalizable no-reference image
  quality assessment via deep meta-learning. IEEE Transactions on Circuits and
  Systems for Video Technology  (2021)

\bibitem{zhu2015learning}
Zhu, J.Y., Krahenbuhl, P., Shechtman, E., Efros, A.A.: Learning a
  discriminative model for the perception of realism in composite images. In:
  Proceedings of the IEEE International Conference on Computer Vision. pp.
  3943--3951 (2015)

\bibitem{zhu2021image}
Zhu, M., He, D., Li, X., Li, C., Li, F., Liu, X., Ding, E., Zhang, Z.: Image
  inpainting by end-to-end cascaded refinement with mask awareness. IEEE
  Transactions on Image Processing  \textbf{30},  4855--4866 (2021)

\end{thebibliography}
